\documentclass[pmlr]{jmlr}

\RequirePackage{graphicx}
 \usepackage{booktabs}
\usepackage{longtable}
\usepackage{float}
\makeatletter
\def\set@curr@file#1{\def\@curr@file{#1}} 
\makeatother
\usepackage[load-configurations=version-1]{siunitx} 
\usepackage{placeins}

\theorembodyfont{\upshape}
\theoremheaderfont{\scshape}
\theorempostheader{:}
\theoremsep{\newline}

\title[Impute With Confidence]{Impute With Confidence: A Framework for Uncertainty Aware Multivariate Time Series Imputation}

\author{\Name{Addison Weatherhead}
       \Email{addison.weatherhead@mail.utoronto.ca}\\ 
       \addr Department of Computer Science\\
       University of Toronto\\
       Toronto, ON, Canada 
       \AND
       \Name{Anna Goldenberg}
       \Email{anna.goldenberg@utoronto.ca}\\ 
       \addr Department of Computer Science\\
       University of Toronto\\
       Toronto, ON, Canada}

\begin{document}

\maketitle

\begin{abstract}
  Time series data with missing values is common across many domains. Healthcare presents special challenges due to prolonged periods of sensor disconnection. In such cases, having a confidence measure for imputed values is critical. Most existing methods either overlook model uncertainty or lack mechanisms to estimate it. To address this gap, we introduce a general framework that quantifies and leverages uncertainty for selective imputation. By focusing on values the model is most confident in, highly unreliable imputations are avoided. Our experiments on multiple EHR datasets, covering diverse types of missingness, demonstrate that selectively imputing less-uncertain values not only reduces imputation errors but also improves downstream tasks. Specifically, we show performance gains in a 24-hour mortality prediction task, underscoring the practical benefit of incorporating uncertainty into time series imputation.
\end{abstract}

\section{Introduction}

Missingness is a common real-world issue found in time series data. An ML practitioner has two choices when confronted with missing values in their data: A) Build their model such that it is aware of the missing values, for example \cite{DBLP:journals/corr/ChePCSL16}, or B) Impute the missing values with their best estimate. Early imputation methods consisted of simple statistics or heuristics, such as imputing with the most common observed value for that variable, or the mean of observed values for that variable. While these are simple and efficient methods, they are typically very inaccurate. More sophisticated and accurate methods have been developed using Machine Learning methods like Linear Regression \cite{mice}, Transformers \cite{saits}, and mixed-method approaches \cite{hyperimpute}.

Missingness in a healthcare setting provides unique challenges. Often times a hospital bed be equipped with various sensors to measure vitals of the patient such as heart rate, blood pressure, blood oxygen saturation, respiratory rate, and so forth. However when patients are taken for a lab test or a short procedure, most or all of these sensors are disconnected for a period of time. Additionally, there can be cases where a sensor disconnects and only comes back online when a doctor or nurse notices and reconnects the sensor. Therefore, in a medical setting there are often cases where the time series data recorded for a given patient has contiguous blocks of missingness, for some or even all of the variables. This poses a unique challenge for imputation and downstream tasks like mortality prediction, which helps motivate this work. 

Lastly, we found there to be an under-representation in time series imputation methods that focus on uncertainty. In our experience, a given imputation method can vary widely in its accuracy, depending on the sample. Having an imputation method that quantifies its uncertainty and makes decisions based on it is an important step towards building more accurate imputation methods and building trust among practitioners in the imputation methods they use. While there have been other approaches for uncertainty informed imputation methods proposed, they typically require integration of a downstream task in order to measure the imputation uncertainty, e.g. \cite{clasfctn_uncrtny_multi_imputed_data}, or limit the number of imputation methods compatible with the uncertainty measure, e.g. \cite{Uncertainty_Aware_VAEs}.

In this work we utilize an uncertainty metric that can apply to a wide class of deep learning imputation methods, without the need for highly specific architecture choices and is not reliant on downstream task performance. We borrow from \cite{dropout_bayes}, which shows that a deep neural net with dropout is "mathematically equivalent to an approximation
to the probabilistic deep Gaussian process", and utilize variance of monte carlo samples from the model output as our uncertainty measure. We empirically find this metric correlates highly with model error, benchmark its ability to capture low and high quality imputations in a variety of simulated and real world Electronic Health Record (EHR) datasets, and evaluate the impact on downstream prediction models. 

\subsection*{Generalizable Insights about Machine Learning in the Context of Healthcare}
Our work shows that there exists a simple and easy uncertainty measure that can be computed for a large class of time series imputation models, and that in some cases choosing not to impute highly uncertain values leads to better downstream task performance. The validity of our uncertainty measure is showcased by its clear, typically linear, relationship with imputation error. Having a simple, widely applicable, and accurate uncertainty measure in time series imputation methods is very useful for models deployed in healthcare settings. Uncertainty measures can be communicated to end users to indicate how certain or uncertain a model is in its predictions. Uncertainty measure can also be incorporated into downstream tasks as input variables, to allow downstream prediction models to utilize imputed values and their accompanying uncertainty values to make more accurate predictions.

\section{Related Work}
Missingness in ML is a long studied problem. The simplest approach to dealing with missing values in a dataset is to delete those records that contain missing values. However, as shown by \cite{LittleRubin2019}, this can lead to bias in analysis if the missing values are not completely randomly distributed (and they are often not). Alternatively, one can replace missing values with the mode, median, or mean of the missing variable, but again this typically leads to biased analysis results. 

Imputation methods utilize information about the observed values and/or patterns of missing values to compute missing values. \cite{mice} developed Multiple Imputation by Chained Equations (MICE), which is a very common approach to imputation. It imputes the dataset multiple times, at each step conditioning on all other values (either observed or imputed) to regress on the missing values of the variable at hand. It repeats this cyclic process a predetermined number of times. More recent approaches to imputation take advantage of deep learning models. 

In the time series domain, there were many recurrent network-based approaches such as \cite{brits}, which uses bidirectional Recurrent Neural Networks (RNNs) to impute. \cite{cdsa} was the first approach to use the self attention based Transformer models for time series imputation. \cite{saits} developed Self-Attention-based Imputation for Time Series (SAITS) which utilizes an augmented Transformer encoder architecture creating both temporal and feature correlations. It has a unique objective function which includes both error on masked out values, as well as error on reconstruction of observed values. IT achieved SOTA results at the time of publishing on multiple datasets.   

There is a class of time series deep learning models that can perform uncertainty estimation. For example, \cite{csdi} developed Conditional Score-based Diffusion models for Imputation (CSDI), which utilizes the denoising process of score-based diffusion models to impute missing data. \cite{gp-vae} developed GP-VAE, a latent variable model which uses a Variational Autoencoder (VAE) and models the lower dimensional representation over time via a Gaussian Process (GP). Recent work by \cite{Uncertainty_Aware_VAEs} incorporated a VAE to capture feature correlations with a recurrent imputation model and include an uncertainty aware mechanism in their imputation model. All these approaches have specific architectural requirements to enable generation of uncertainty values for their imputations (i.e. VAE-based, diffusion-based, etc). The field of Multivariate Time Series imputation methods is vast, we mention several key methods in this work, but for brevity are unable to mention all of the relevant and important work done in this field. 

Lastly, \citep{dropout_bayes} is a relevant work that formulates the primary uncertainty measure we use. In their work, they show that dropout can be interpreted as performing approximate Bayesian Inference in a Deep Gaussian Process. They introduce Monte Carlo Dropout -- essentially sampling from the model's predictive distribution by conducting multiple forward passes with dropout turned on. They utilize the variance of the Monte Carlo samples as an uncertainty measure and showcase its effectiveness.

\section{Methods}
\subsection{Notation}
We denote a sample of time series data $X_{obs} \in \mathbb{R}^{T 
\times D}$, where $T$ is the number of time steps and $D$ is the dimensionality of the data. It may contain missing values, which we call \emph{genuine missingness}, so alongside $X_{obs}$ we have $M_{obs} \in \{0,1\}^{T \times D}$, a binary mask indicating which values are observed ($0$) and which are missing ($1$). The 'obs' indicates that this is the observed data, straight from the dataset, with no alterations. During the training of a deep learning-based imputation method for time series data, manual masking needs to be done on some non-missing values in order to have a 'missing' value for the model to impute and a ground truth target for loss calculation. We denote $X_{cor} \in \mathbb{R}^{T 
\times D}$ as the sample $X_{obs}$ that has been corrupted with some form of \emph{synthetic missingness} (see Sec \ref{types_of_missingness}). Note $X_{cor}$ contains both genuinely missing values and values that have been synthetically masked out. 

\subsection{Types of Missingness} \label{types_of_missingness}

In our work, we utilize multiple patterns of synthetic missingness. These are used for training, validation and testing. We outline the types of missingness below and discuss how we practically create that type of missingness in our data. 

\begin{itemize}
    \item Missing Completely At Random (MCAR): This type of missingness assumes that the mask of missing values in no way relies on the data itself \citep{missingness_survey}. In practice, we create MCAR missingness by selecting random values (uniformly across both dimensions and time) to mask out the values. 
    \item Missing At Random (MAR): This type of missingness assumes that the probability of a value being missing is defined by the observed variables \citep{missingness_survey}. Following \cite{yaib}, we define a subset of variables to remain fully observed, then run a logistic regression model (with randomly initialized weights) that uses those observed variables to predict the probability of missingness in the remaining columns. The intercept of this logistic model is chosen to achieve a target 30\% missing rate. Because the missingness depends solely on variables that remain observed, once you condition on these observed variables, the mechanism is MAR.(Note: For the purpose of the LR model, any genuine missing variables in $X_{obs}$ are replaced with the mean of that variable. This is strictly for the LR to be run and is not used henceforth).
    \item Missing Not At Random (MNAR): The missingness mask is neither MCAR nor MAR, it assumes the probability of a value being missing relies both on observed and unobserved values \citep{missingness_survey}. Following \cite{yaib}, 
    we split our variables into a set used by a logistic regression model (the ‘predictors’) and a set that is masked according to that model’s output (the ‘targets’). We also mask some of the predictor set completely at random (MCAR). This ensures that missingness in the target columns depends partly on variables that are themselves unobserved, producing MNAR behavior. We choose the logistic intercept to keep the overall missingness at 30\%. (Note: Again, for the purpose of the LR model any input variables that are genuinely missing in $X_{obs}$ we fill with the mean of that variable, just for the purpose of running the LR model. This mean is not persisted into the data).
    \item Black Out (BO): This form of missingness is not standard in missingness literature. It was created in the Yet Another Icu Benchmark (YAIB) package by \cite{yaib} to create missingness across all variables. This form of missingness sets all variables at random time steps to missing so as to achieve the desired level of missingness ($30\%$)
    \item Block Black Out (blockBO): We implement this form of missingness to simulate what we have observed in the real world ICU time series data -- where multiple or all variables go missing for a contiguous block of time due to a sensor disconnection or the patient being taken to a procedure. As implemented, blockBO inserts contiguous time blocks of missingness across all variables. Blocks range in size and care is taken to ensure blocks do not overlap. Blocks are inserted until the desired level of missingness ($30\%$) is achieved.
\end{itemize}

\subsection{Uncertainty Estimation}
Following \cite{dropout_bayes}, we recognize that a deep neural network with dropout layers after each weight layer is a Bayesian approximation of a deep Gaussian Process (GP). Their work shows that the first and second moments of the model's output distribution can be derived simply as the sample mean and sample variance, respectively, of $F$ stochastic forward passes of the model. We follow this same structure; for a given sample $X_{cor}$ we compute $U_{cor} \in \mathbb{R}^{T \times D}$, which is the elementwise standard deviation of $F$ forward passes of the imputation model (we use standard deviation as its values are slightly more interpretable). Similarly, we can use the same process to create a $U_{obs}$ for $X_{obs}$ with genuine missingness, which can be useful at inference time.

\subsection{Calibration on a Validation Set} \label{calibration_step}

In order to assess the tradeoff between model uncertainty and imputation error, as well as the effects of selective imputation on downstream tasks, we develop a calibration step to select a number of uncertainty thresholds to experiment with. After training an imputation model using one of the masking methods discussed in Section \ref{types_of_missingness}, we compute various quantiles of the distribution of uncertainty values on the validation set only (using only values that were synthetically masked), and use the quantile boundaries as threshold values. Concretely, we choose $t_{10}$, $t_{20}$, ..., $t_{100}$ as our thresholds, representing the $10\%$, $20\%$, ..., $100\%$ quantiles of the distribution $\mathcal{U} = \{U_{cor}^{(n)}(t,d) \vert (M_{cor}^{(n)}-M_{obs}^{(n)})=1; n=1,...,N, t=1,...,T, d=1,...,D\}$, where $N$ is the number of validation samples and $U_{cor}^{(n)}(t,d)$ is the computed uncertainty for a corrupted sample at time step $t$ and variable dimension $d$. Intuitively, $\mathcal{U}$ is computing uncertainty values across all variables/timesteps, and keeping only the uncertainty values for timestep/variable combinations that were artificially masked (this is where the $(M_{cor}^{(n)}-M_{obs}^{(n)})=1$ part comes from). We compute thresholds $t_10$, $t_20$, ..., $t_100$ for the purpose of \emph{selective imputation}, this is the practice of taking a simple with genuine missing values and imputing only those where the model's uncertainty is below a given threshold. Note that $t_{100}$ is chosen so that selectively imputing values with uncertainty $<=t_{100}$ effectively yields full imputation. From there, we conduct experiments studying the validation imputation error rate as well as downstream task performance when selectively imputing values with uncertainty less than or equal to each threshold. Experimental results showcase the efficacy of the uncertainty measure and highlight that in some cases, selectively imputing only values where the imputation model is sufficiently confident yields stronger downstream task performance.

\section{Data}

We use the following ICU datasets for our experiments: 

\begin{itemize}
    \item MIMIC-IV \citep{mimiciv_dataset}: A large, publicly available dataset containing de-identified electronic health records from patients admitted via the Emergency Department or cared for in the ICU, including patient demographics, vital signs, laboratory tests, and clinical notes. Data procured from the Beth Israel Deaconess Medical Center in Boston, MA.
    \item eICU \citep{eicu_dataset}: A multi-center intensive care unit (ICU) dataset that includes data from many hospitals across the US, capturing patient admissions, vital signs, treatment interventions, and outcomes.
    \item HiRID \citep{hirid_dataset}: A high-resolution ICU dataset containing detailed physiological signals and clinical data relating to over $30,000$ patients from the Department of Intensive Care Medicine of the Bern University Hospital, Switzerland.
\end{itemize}

The processing of each ICU dataset is the same. Data is split into \emph{static variables} (things like age, weight, etc) and \emph{dynamic variables} (things like heart rate, blood pressure, etc). Following the data extraction pipeline used by \cite{yaib}, we remove any patients with negative end time stamps (these are corrupted/invalid samples), we remove patients whose Length Of Stay (LOS) is $< 30$ hours, we remove patients who died within the first $30$ hours of admission, we remove patients who have less than 4 recorded measurements for their entire stay, we remove patients with over 12 hours of contiguous missingness for any dynamic variable, and we remove any patient less than $18$ years of age. We save the first 24 hours of data for each patient (at an hourly frequency), and for the prediction task of mortality a patient is labeled with $1$ if they die during the visit and $0$ otherwise. After all patient filtering has been done the remaining number of patients are $12,859$ for HiRID, $113,381$ for eICU, and $49,523$ for MIMIC IV.

For our imputation and classification experiments, we utilize the 6 most observed dynamic variables in the dataset: Heart Rate, Mean Arterial Pressure, Systolic Blood Pressure, Diastolic Blood Pressure, Respiratory Rate, and Blood Oxygen Saturation. For the classification experiments only we also include the static variables as input to the classification model; Age, Sex, Height, and Weight. Here the input variables also include partially imputed dynamic variables, along with a missing indicator for each variable at each time step, historical (up to the current time step) min, max, and mean for each variable, and lastly a historical counter which counts the number of missing values up to the current time step for each variable. We split each dataset into train ($40\%$), validation ($10\%$), and test sets ($50\%$). Each dataset is standardized for each split.


\section{Experimental Results} 
\subsection{Model Choice}
We train a SAITS model and a Transformer encoder model (with an extra linear layer to project the encodings to the dimension of the data). These models are chosen because they are among the best performing imputation methods for EHR datasets as shown by \cite{imputation_benchmarks}, we reproduce their benchmark comparison on several methods, and also confirm SAITS and the Transformer encoder as the best performers (This comparison can be found in Appendix \ref{AppendixB}). Additionally, they have a dropout layer between each linear layer, making them suitable candidates for the uncertainty measure we use. For any given training run, we choose one of the missingness types detailed in Section \ref{types_of_missingness} to create synthetic missingness in the training set to train the imputation model on. We use the same missingness type for validation and testing performance as well. We perform a grid search for each combination of model, dataset, and missingness type according to Table \ref{tab:hp_grid}, and select the best model of each sweep based on validation set Mean Absolute Error (MAE).

\begin{figure*}[h]
    \centering
    \includegraphics[width=0.45\linewidth]{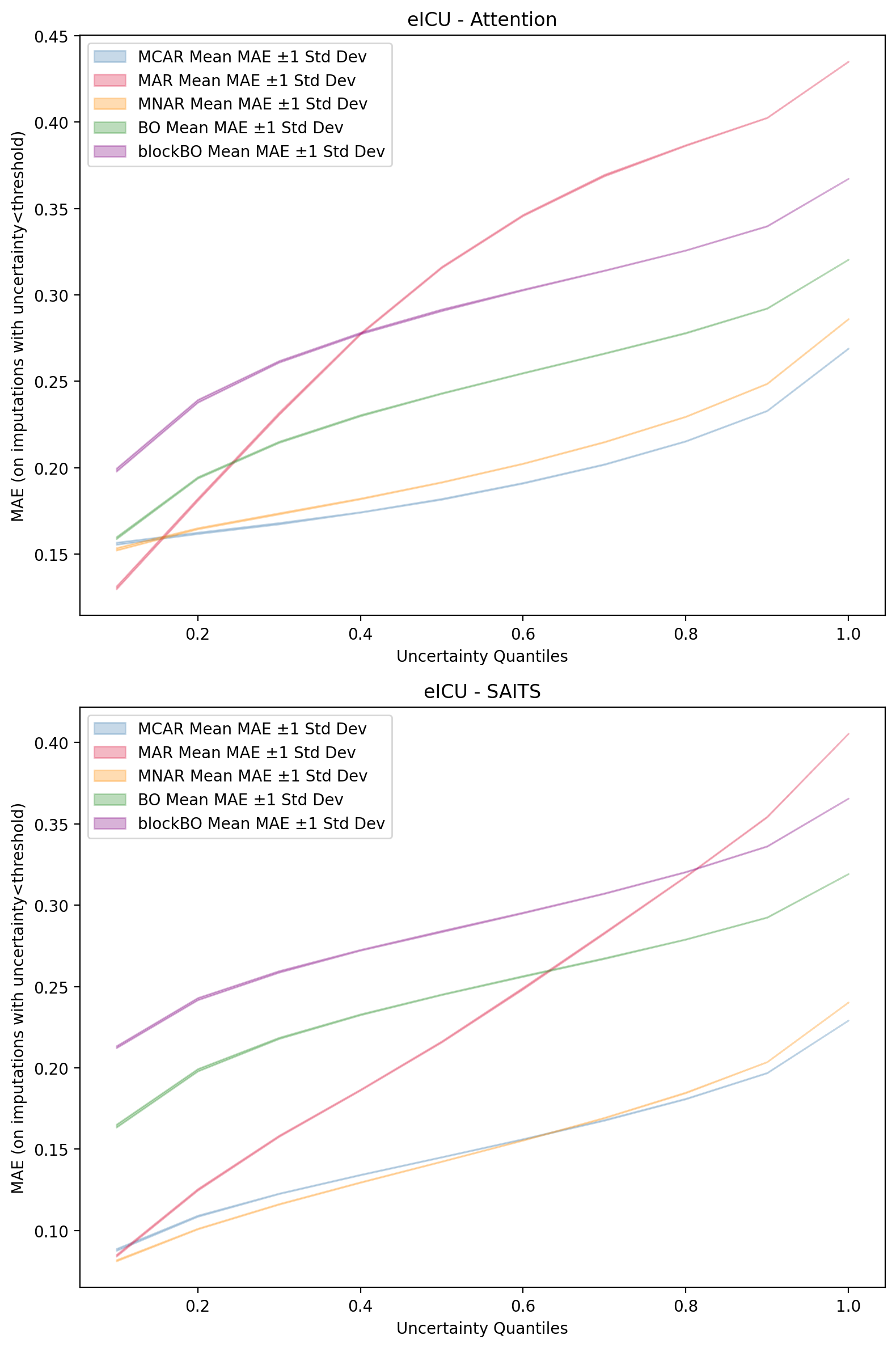}
    \includegraphics[width=0.45\linewidth]{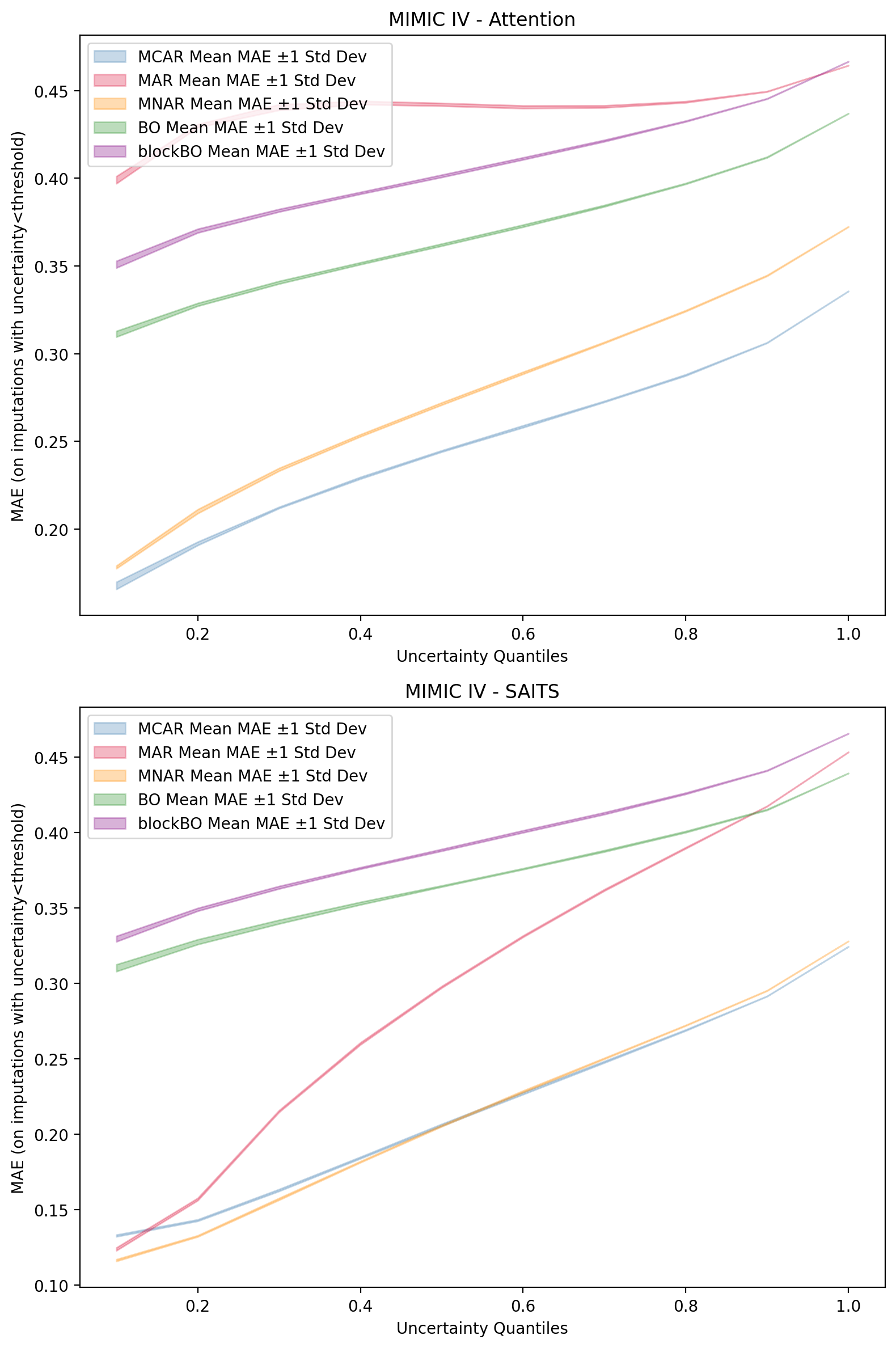}
    \caption{Impact of Uncertainty on MAE for eICU (left) and MIMIC IV (right).}
    \label{fig:eicu_miiv_uncertainty}
\end{figure*}

\subsection{Uncertainty Evaluation} 
\begin{figure}[!h]
   \centering 
   \includegraphics[width=3in]{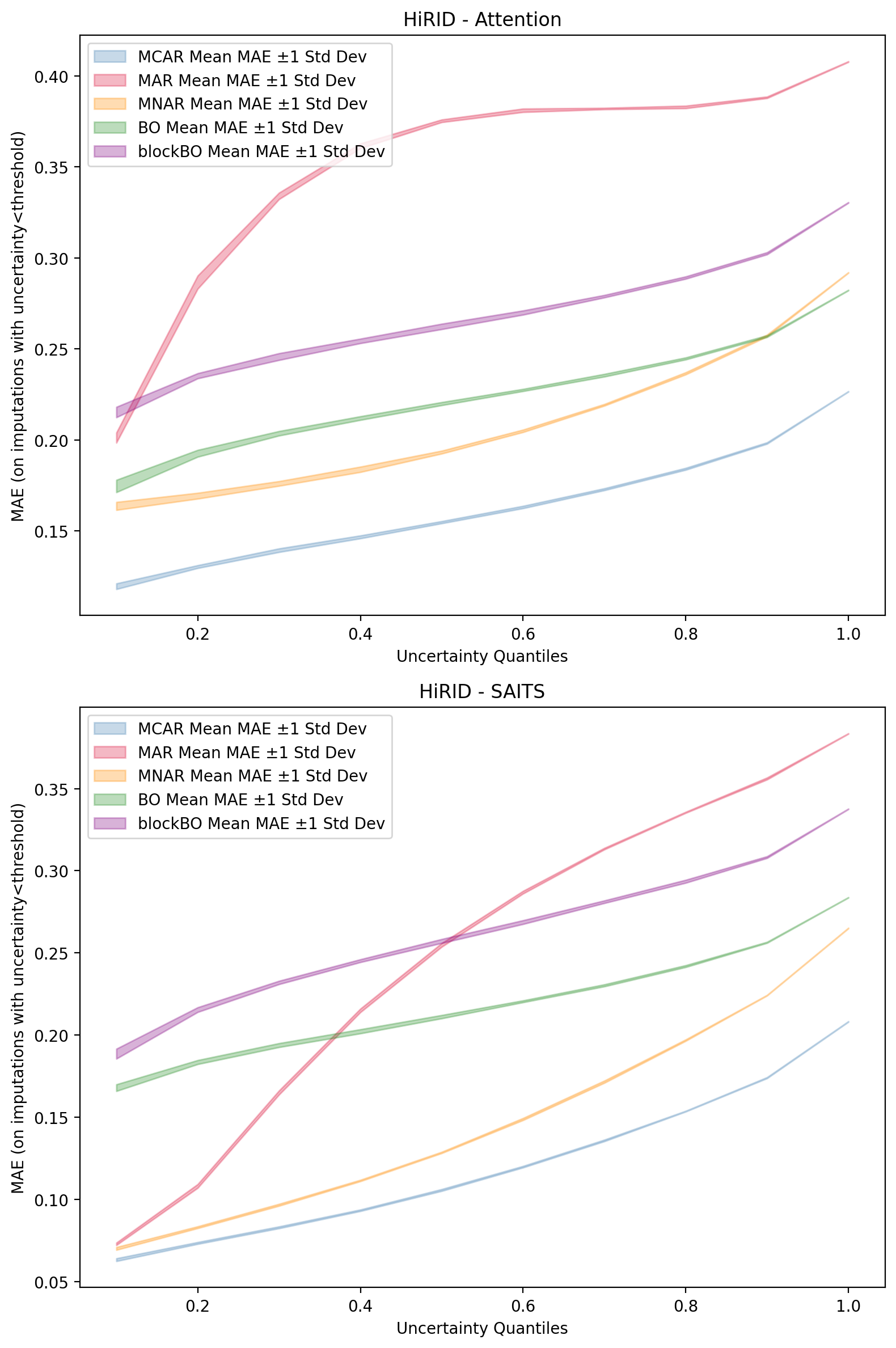} 
   \caption{Impact of Uncertainty on MAE for HiRID}
   \label{fig:hirid_uncertainty} 
 \end{figure}

First, we evaluate the validity of the uncertainty measure and its relationship with imputation error. We do so by following the calibration step outlined in Section \ref{calibration_step}. We use $M=16$ Monte Carlo samples from each model's output with dropout turned on. As discussed earlier, the standard deviation of the model's predictive distribution is used as our uncertainty measure. Once thresholds $t_{10}, t_{20}, ..., t_{100}$ have been obtained, we plot the MAE on the data imputed for values with uncertainty less than or equal to each threshold. Note the MAE here is normalized only for the values we actually impute. As we can see in Figure \ref{fig:eicu_miiv_uncertainty}, we plot the relationship between model uncertainty and MAE for eICU and MIMIC IV across all missingness types and for each model architecture we train. The corresponding plots for HiRID can be found in Figure \ref{fig:hirid_uncertainty}. In general, we can see a very clear and consistent trend across all datasets, both models, and all missingness types that the higher level of uncertainty we allow in model imputations, the higher absolute error it makes in those imputations. This validates that the uncertainty measure we use is accurate at actually capturing model confidence. It is also important to note that all data is standardized, so some of the changes we see in absolute error imputation performance across different thresholds is quite large. For example, the Transformer with MAR missingness on eICU shows a change of $0.3$ in MAE across the different levels of uncertainty. Of course there is a tradeoff here, as we decrease the amount of allowable model uncertainty, we impute fewer values and leave more as missing. To investigate this further, we run experiments testing its impact on downstream task performance.

\subsection{Downstream Task Impact}
\begin{figure}[h]
   \centering    
   \includegraphics[width=3in]{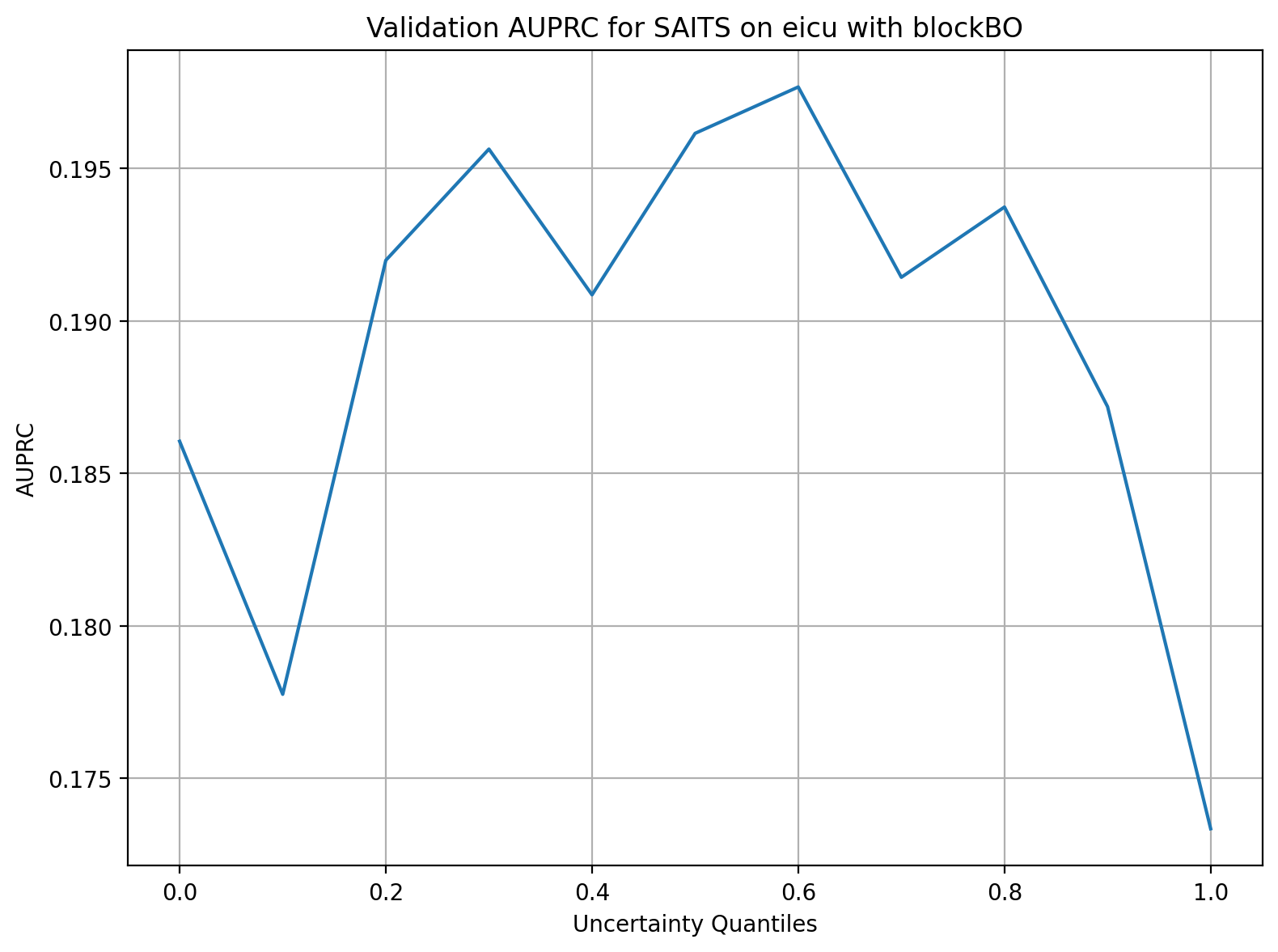} 
   \caption{Impact of Uncertainty on AUPRC for a classifier trained for mortality prediction, using a pretrained imputation model trained for blockBO missingness on eICU}
   \label{fig:classfcn_eicu_SAITS_blockBO} 
 \end{figure}

\begin{figure}[h]
   \centering 
   \includegraphics[width=3in]{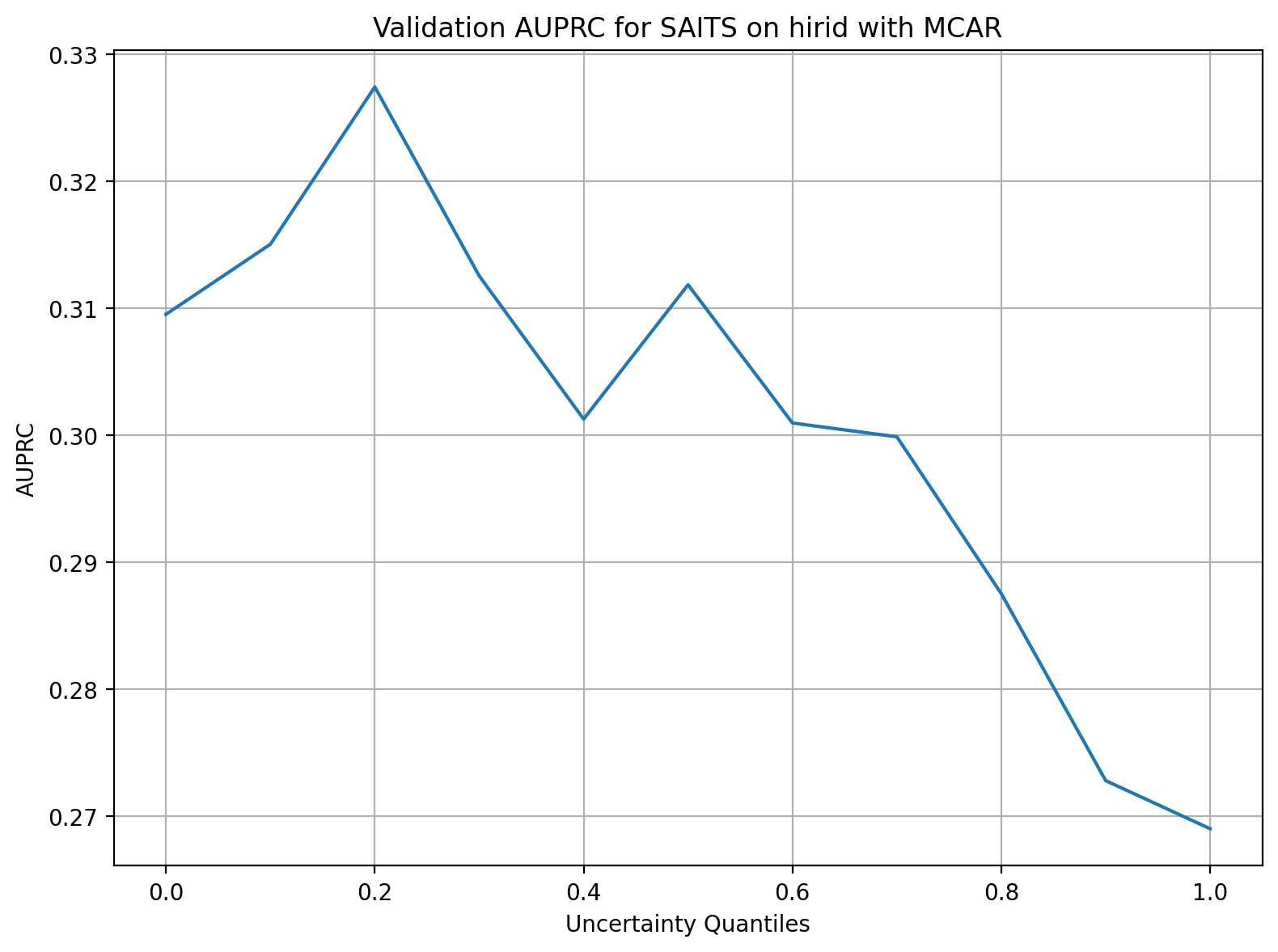} 
   \caption{Impact of Uncertainty on AUPRC for a classifier trained for mortality prediction, using a pretrained imputation model trained for MCAR missingness on HiRID}
   \label{fig:classfcn_hirid_SAITS_MCAR} 
 \end{figure}

We utilize the thresholds $t_{10}, t_{20}, ..., t_{100}$ as well as an initial $t_0$, which is equivalent to not imputing at all. For each threshold, we selectively impute according to that threshold (note imputation for the classification task is performed on $X_{obs}$ samples with no synthetic missing values) using the now pre-trained imputation model, then train and test a Light Gradient-Boosting Machine (LGBM) \cite{lgbm} mortality classifier. From there we obtain validation set AUPRC to measure model performance. In Figure \ref{fig:classfcn_eicu_SAITS_blockBO} we can see a clear case of the downstream task performing better when imputation is only performed on the $60\%$ most confident missing values. When we allow imputation model uncertainty above that, performance starts to degrade. Additionally, we can see on the left at $0.0$, that not imputing at all also degrades performance compared to selective imputation. Conversely, we can see in Figure \ref{fig:classfcn_hirid_SAITS_MCAR} that the ideal threshold for selective imputation for HiRID using a SAITS imputation model trained for MCAR is much smaller, at $0.2$. 

Not all such plots showing downstream performance for each model/missingness type/dataset combination show such clarity on what threshold is ideal for selective imputation. This may indicate that the ones that do have an imputation model trained using a missingness type most similar to the genuine missingness found in the data. The full set of Uncertainty AUPRC plots can be found in the Appendix \ref{Appendix}.

\section{Discussion} 

This work highlights the importance of uncertainty estimation towards thoughtful imputation for real world time series datasets. We demonstrate that in certain cases the downstream model performs better when only selective imputation is applied. This is an important implication for those deploying imputation methods in healthcare settings, as further analysis might reveal the default behavior of imputing every missing value is not the best choice. We show that Monte Carlo Dropout is a powerful tool for deep learning uncertainty estimation. This widely applicable uncertainty measure highly correlates with imputation model error, allowing it to be used for selective imputation, communication to end users, and downstream tasks. 
\paragraph{Limitations}

One notable limitation to this work is that it requires $F$ forward passes to compute the uncertainty measure. At inference time in a limited compute environment, this may be prohibitively expensive. Future work could apply amortization techniques or study the stability of the uncertainty value as one decreases $F$. Additionally, we evaluate our models on a very common but only one task - mortality prediction. Future work could expand this to evaluate the impact of imputation methods on other downstream tasks. Lastly, if the data distribution shifts over time then retraining and recalibration of the imputation method would be required on a regular basis to ensure the model can impute well and the threshold being chosen for selective imputation is reasonable.


\FloatBarrier
\bibliography{sample}

\newpage
\appendix
\section{Additional Tables and Plots}
\label{AppendixA}
Additional Tables and Plots

  \begin{table}[h]
  \centering 
  \caption{Hyperparameters we sweep over in a randomized grid search for each combination of model, dataset, and missingness type.}
  \begin{tabular}{ll}
  \toprule
    \textbf{Hyperparameter} & \textbf{Values} \\
    \midrule
    Learning Rate Scheduler & ["cosine", None] \\ 
    Learning Rate & [0.001, 0.005] \\ 
    Number of layers & [2, 4] \\
    d\_model & [64, 128] \\
    d\_inner & [64, 128] \\ 
    Number of heads & [4, 8] \\
    \bottomrule
  \end{tabular}
  \label{tab:hp_grid} 
\end{table}

\begin{figure}[h]
   \centering    
   \includegraphics[width=3in]{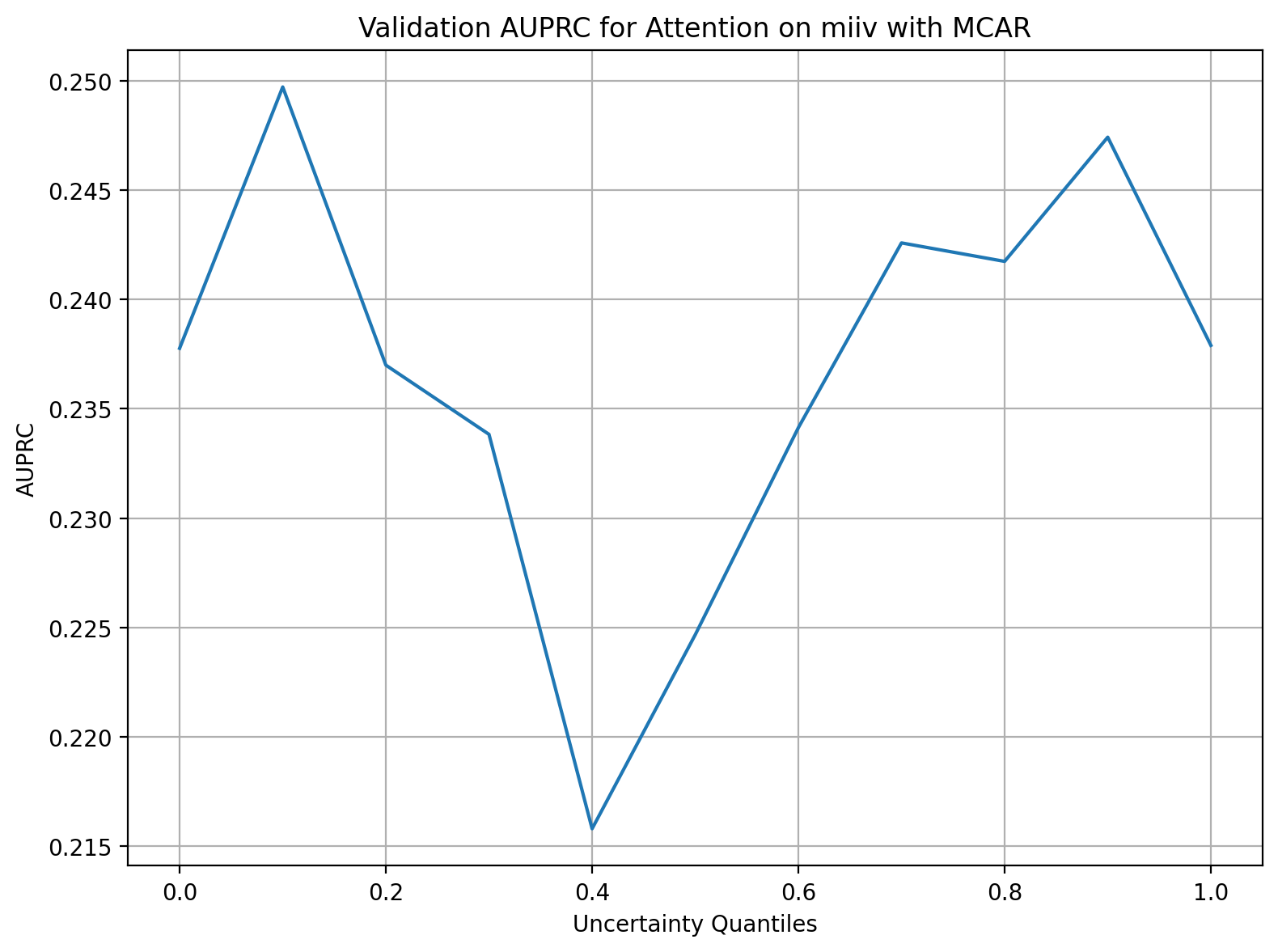} 
   \caption{Impact of Uncertainty on AUPRC for a classifier trained for mortality prediction, using a pretrained imputation model trained for MCAR missingness on miiv}
   \label{fig:fig:classfcn_miiv_Attention_MCAR} 
 \end{figure}

\begin{figure}[h]
   \centering    
   \includegraphics[width=3in]{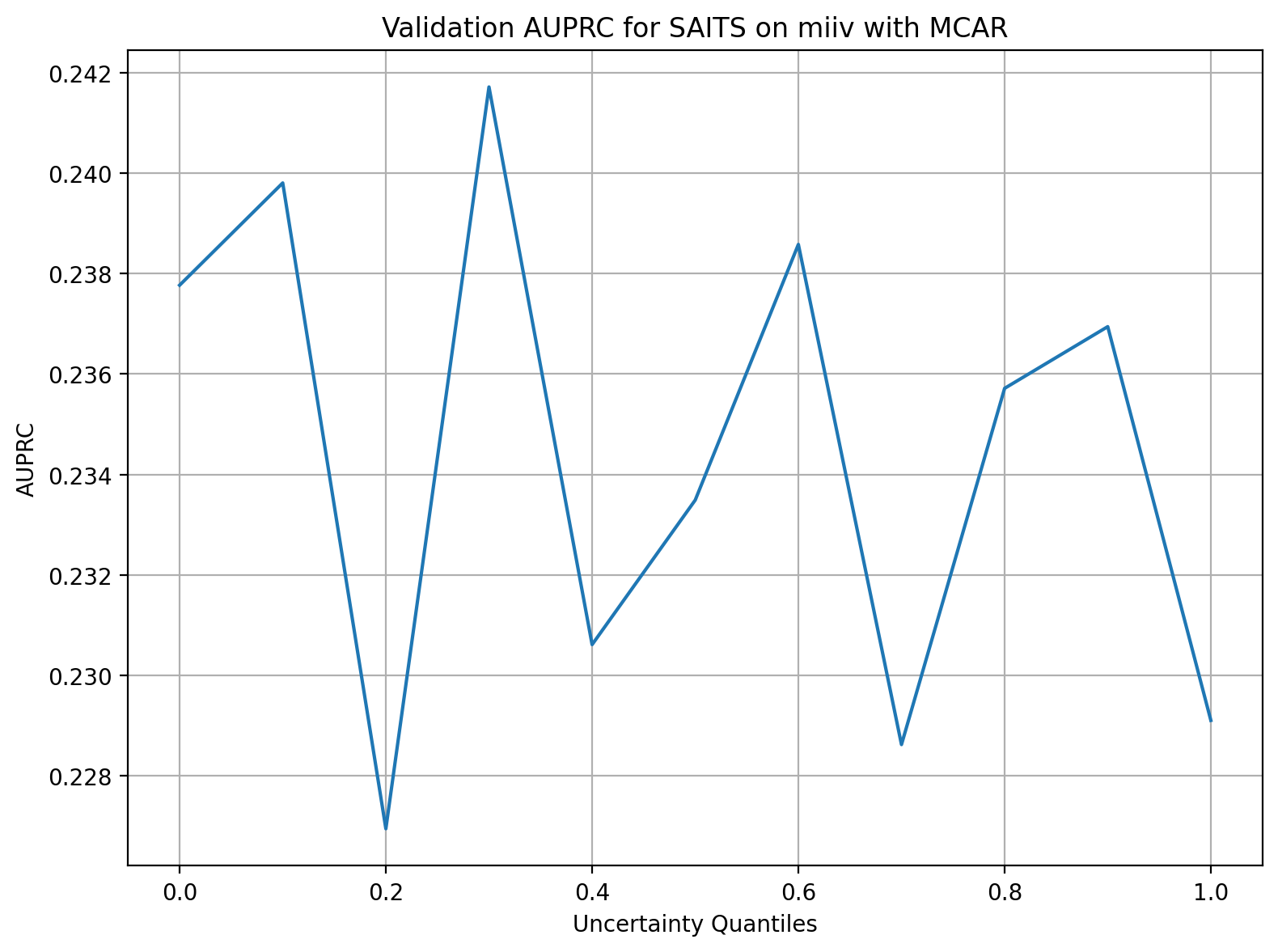} 
   \caption{Impact of Uncertainty on AUPRC for a classifier trained for mortality prediction, using a pretrained imputation model trained for MCAR missingness on miiv}
   \label{fig:fig:classfcn_miiv_SAITS_MCAR} 
 \end{figure}

\begin{figure}[h]
   \centering    
   \includegraphics[width=3in]{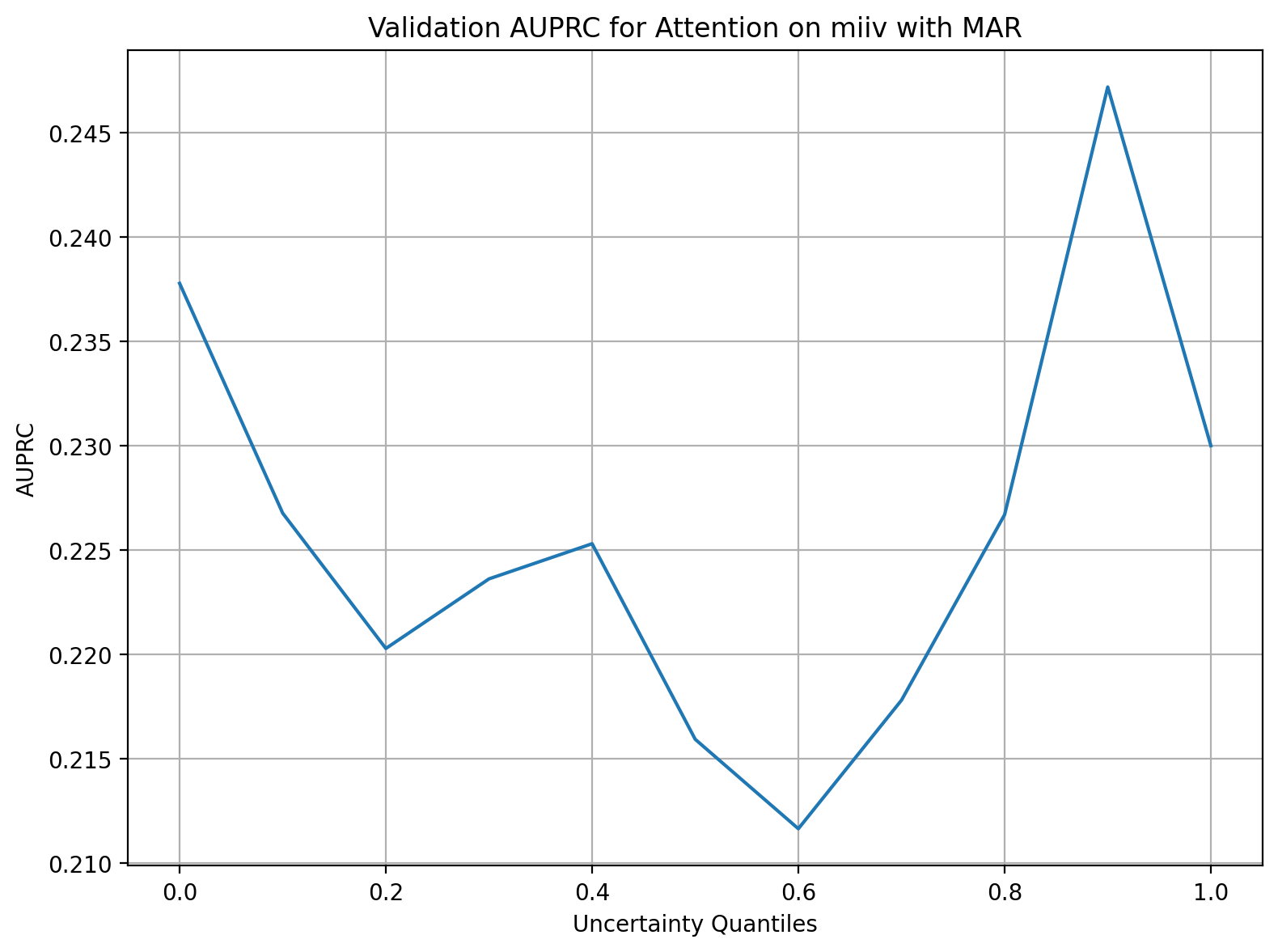} 
   \caption{Impact of Uncertainty on AUPRC for a classifier trained for mortality prediction, using a pretrained imputation model trained for MAR missingness on miiv}
   \label{fig:fig:classfcn_miiv_Attention_MAR} 
 \end{figure}

\begin{figure}[h]
   \centering    
   \includegraphics[width=3in]{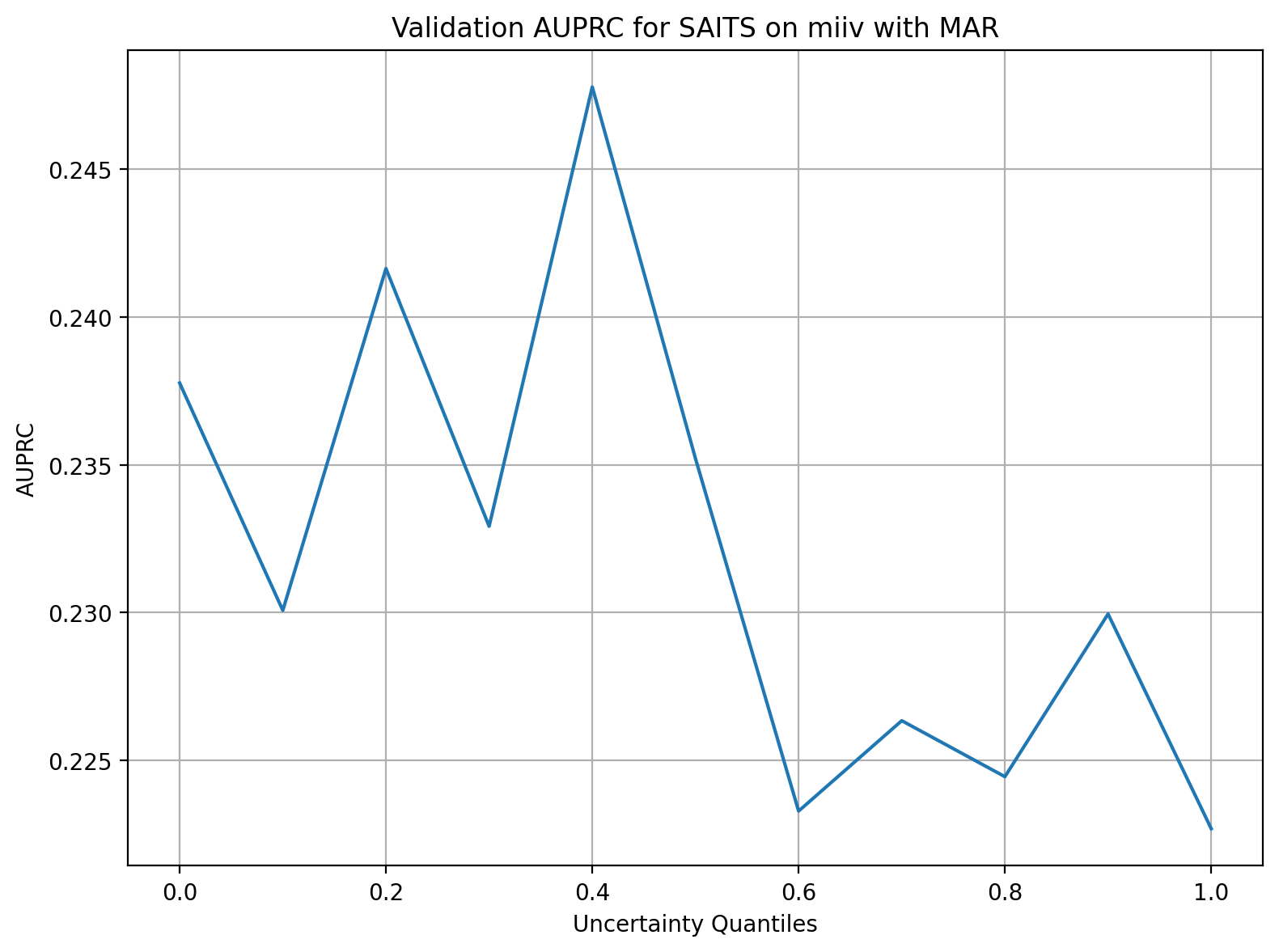} 
   \caption{Impact of Uncertainty on AUPRC for a classifier trained for mortality prediction, using a pretrained imputation model trained for MAR missingness on miiv}
   \label{fig:fig:classfcn_miiv_SAITS_MAR} 
 \end{figure}

\begin{figure}[h]
   \centering    
   \includegraphics[width=3in]{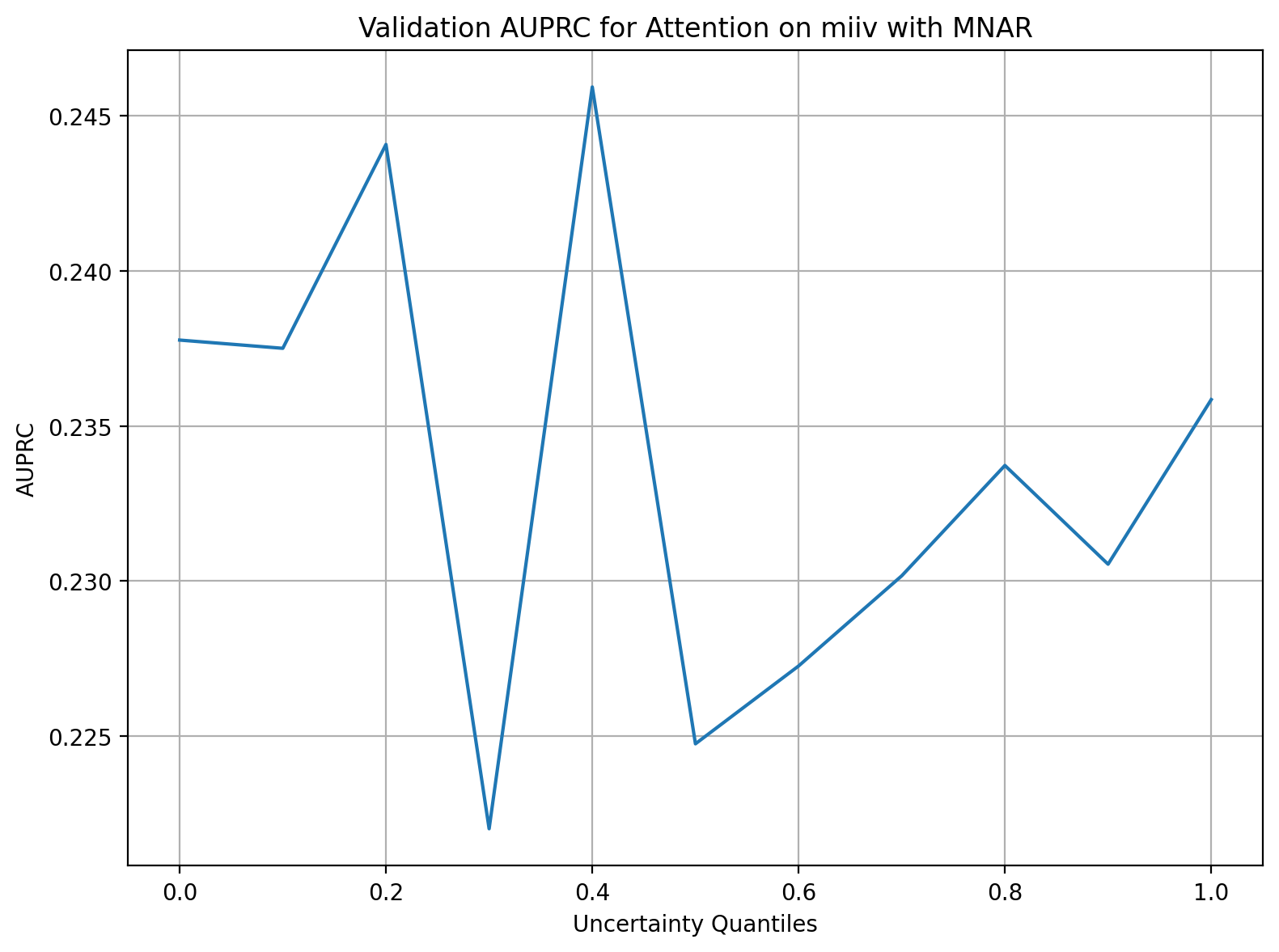} 
   \caption{Impact of Uncertainty on AUPRC for a classifier trained for mortality prediction, using a pretrained imputation model trained for MNAR missingness on miiv}
   \label{fig:fig:classfcn_miiv_Attention_MNAR} 
 \end{figure}

\begin{figure}[h]
   \centering    
   \includegraphics[width=3in]{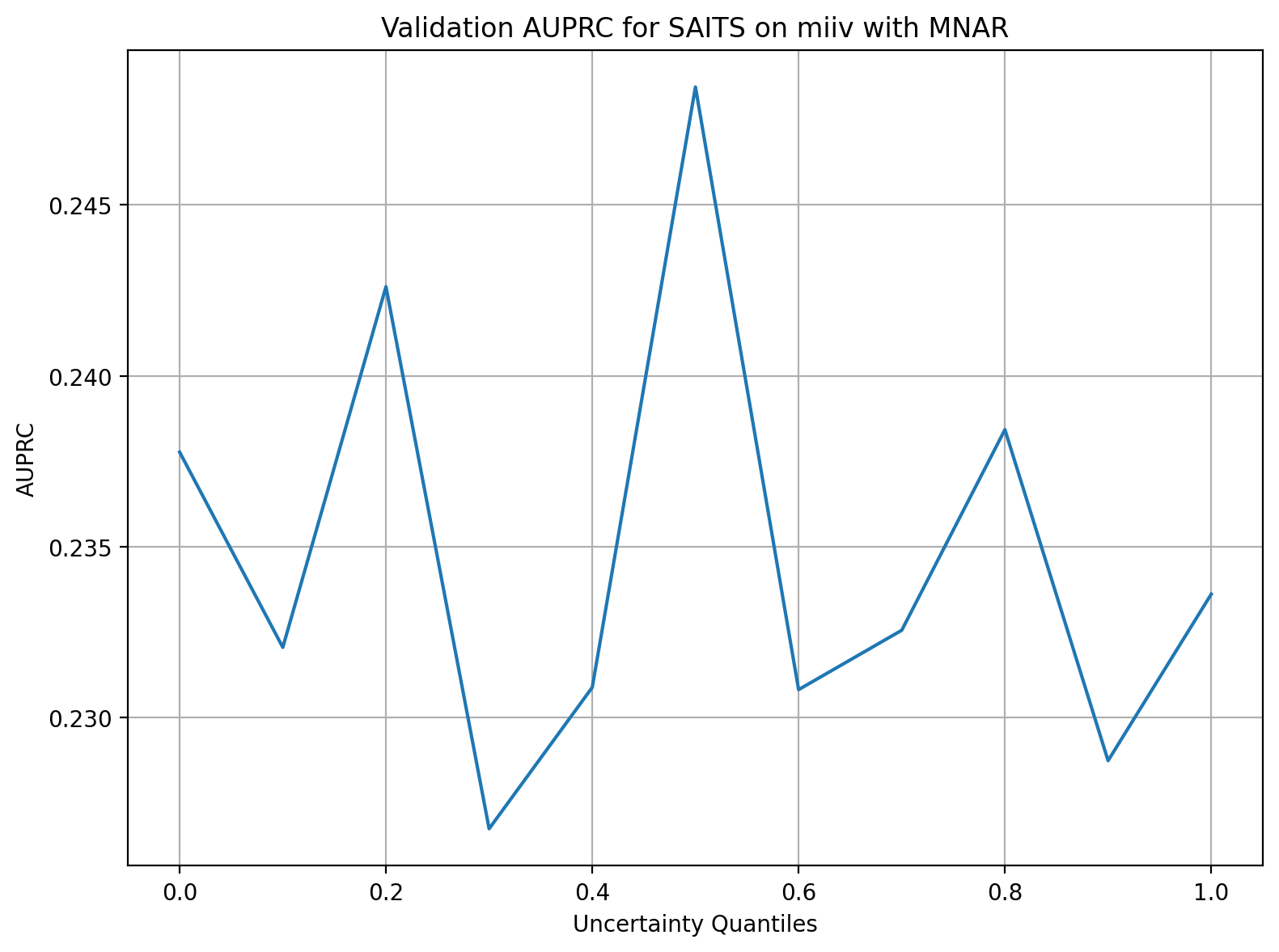} 
   \caption{Impact of Uncertainty on AUPRC for a classifier trained for mortality prediction, using a pretrained imputation model trained for MNAR missingness on miiv}
   \label{fig:fig:classfcn_miiv_SAITS_MNAR} 
 \end{figure}

\begin{figure}[h]
   \centering    
   \includegraphics[width=3in]{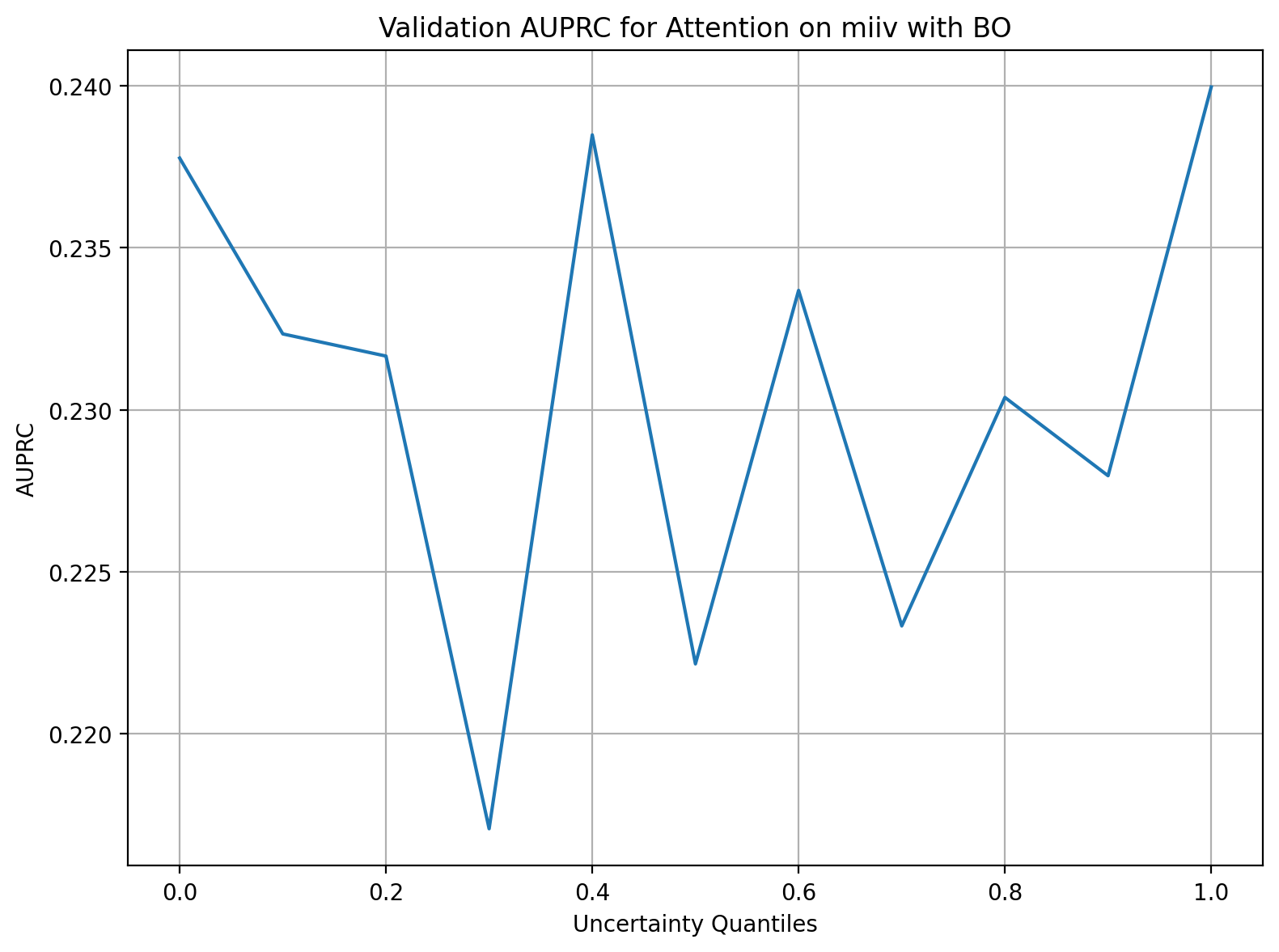} 
   \caption{Impact of Uncertainty on AUPRC for a classifier trained for mortality prediction, using a pretrained imputation model trained for BO missingness on miiv}
   \label{fig:fig:classfcn_miiv_Attention_BO} 
 \end{figure}

\begin{figure}[h]
   \centering    
   \includegraphics[width=3in]{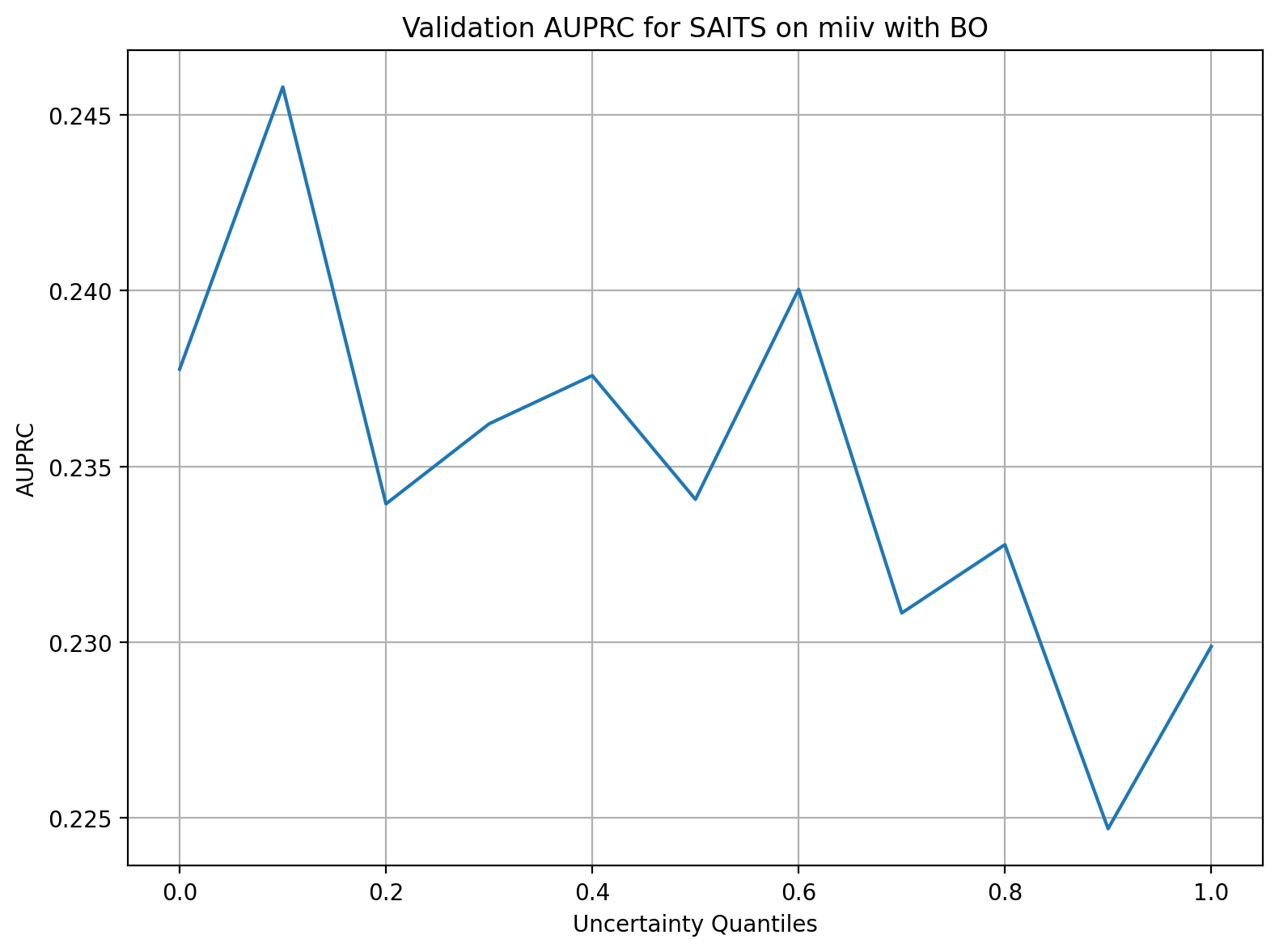} 
   \caption{Impact of Uncertainty on AUPRC for a classifier trained for mortality prediction, using a pretrained imputation model trained for BO missingness on miiv}
   \label{fig:fig:classfcn_miiv_SAITS_BO} 
 \end{figure}

\begin{figure}[h]
   \centering    
   \includegraphics[width=3in]{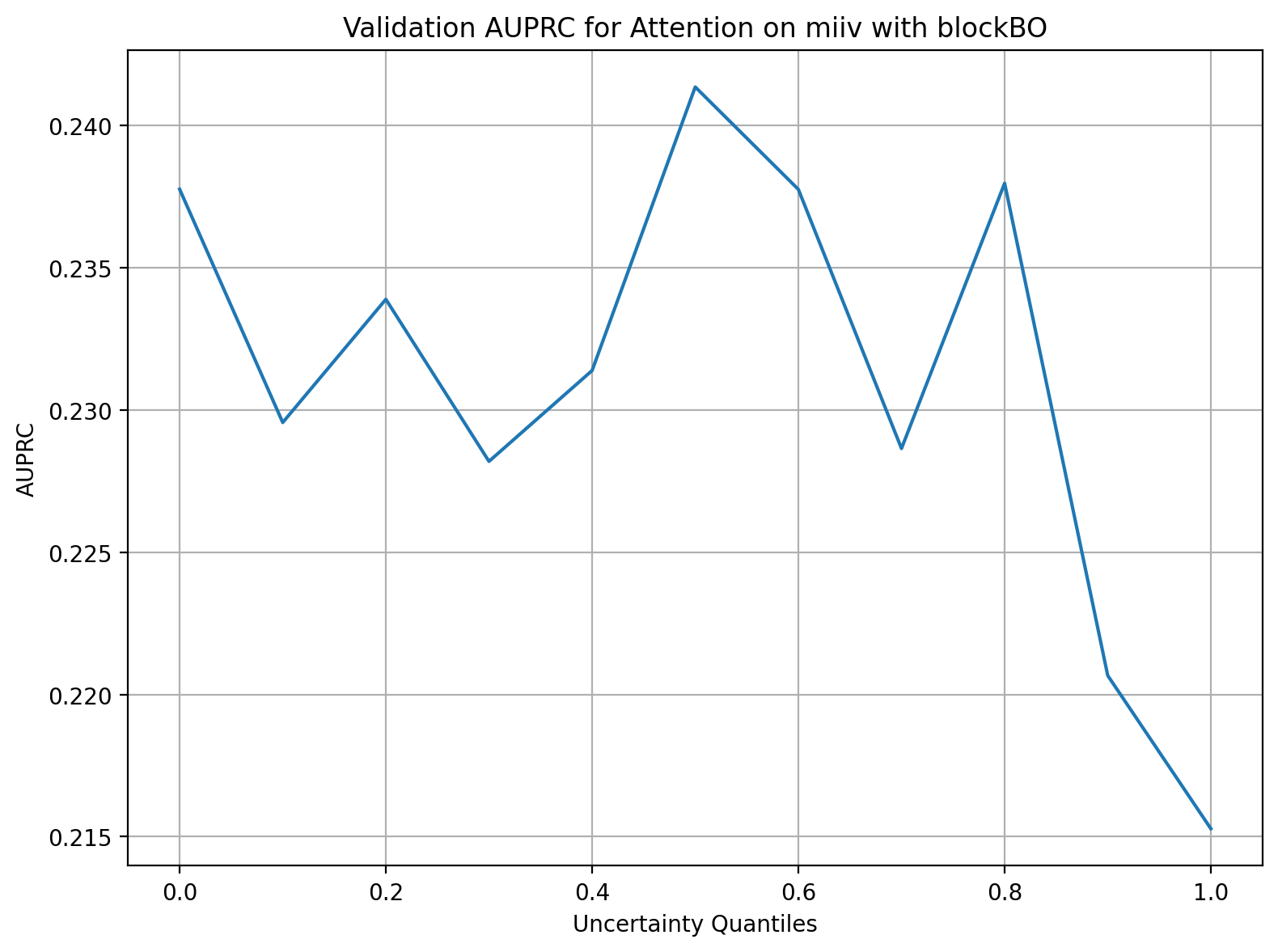} 
   \caption{Impact of Uncertainty on AUPRC for a classifier trained for mortality prediction, using a pretrained imputation model trained for blockBO missingness on miiv}
   \label{fig:fig:classfcn_miiv_Attention_blockBO} 
 \end{figure}

\begin{figure}[h]
   \centering    
   \includegraphics[width=3in]{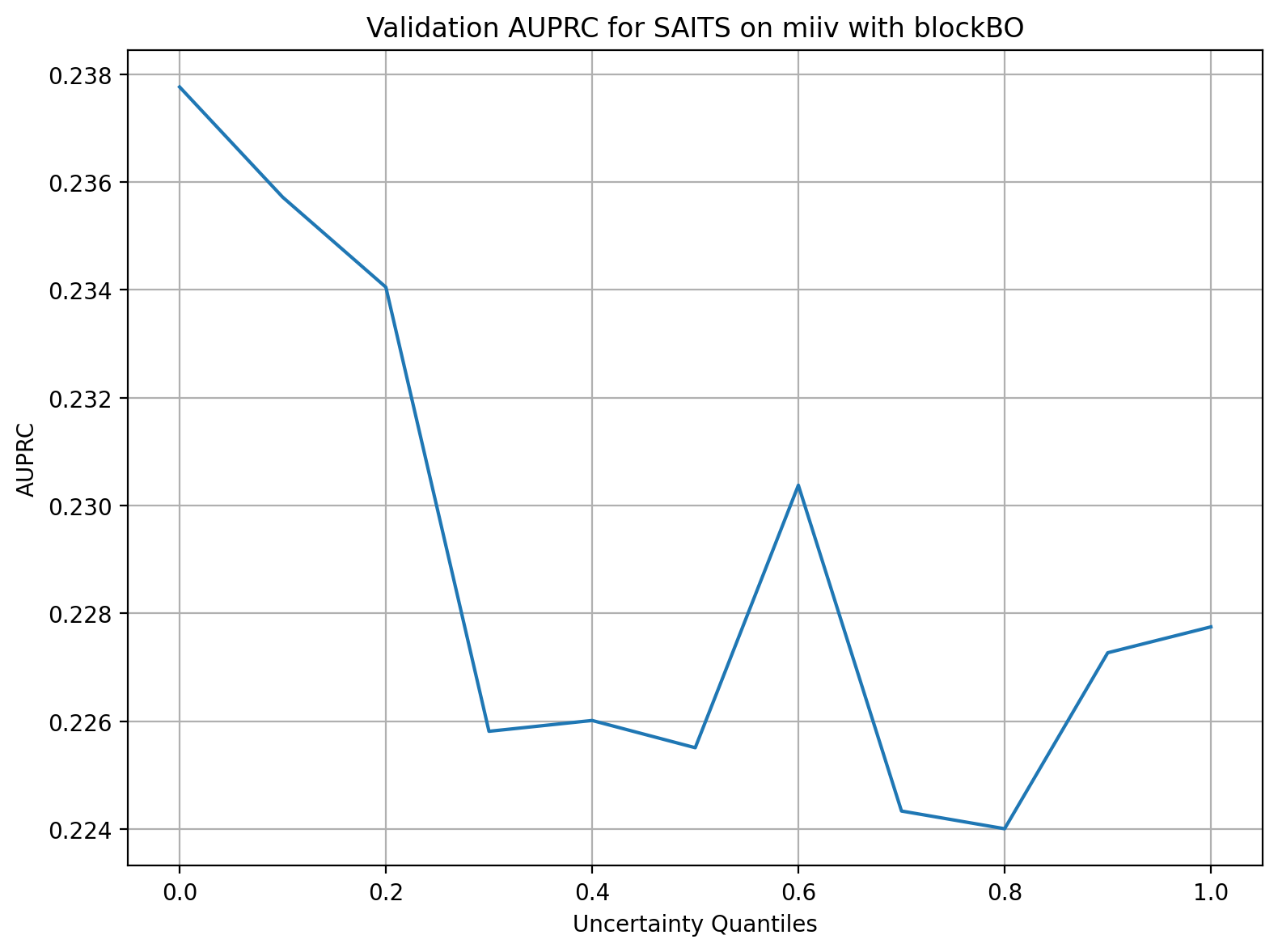} 
   \caption{Impact of Uncertainty on AUPRC for a classifier trained for mortality prediction, using a pretrained imputation model trained for blockBO missingness on miiv}
   \label{fig:fig:classfcn_miiv_SAITS_blockBO} 
 \end{figure}

\begin{figure}[h]
   \centering    
   \includegraphics[width=3in]{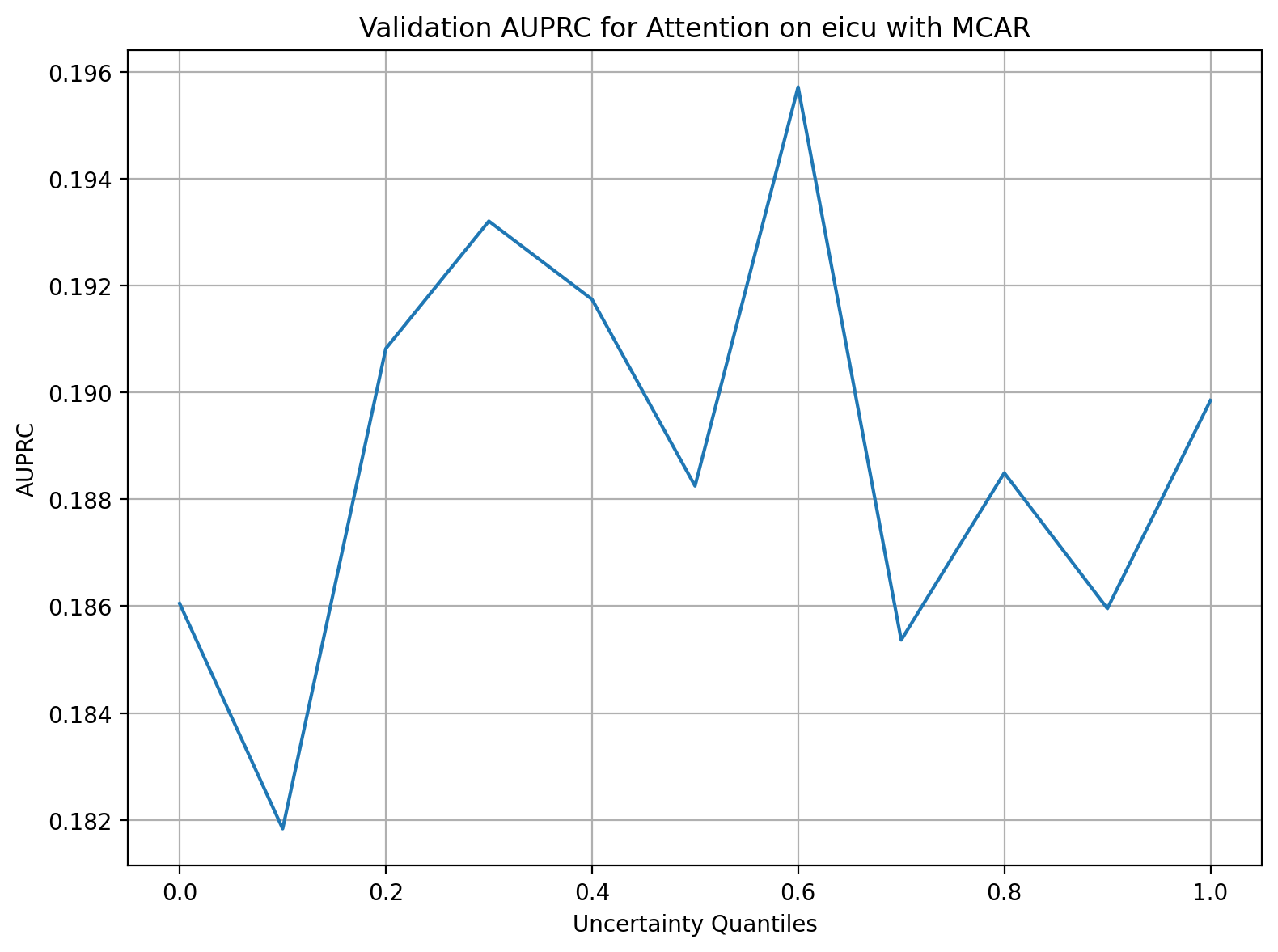} 
   \caption{Impact of Uncertainty on AUPRC for a classifier trained for mortality prediction, using a pretrained imputation model trained for MCAR missingness on eicu}
   \label{fig:fig:classfcn_eicu_Attention_MCAR} 
 \end{figure}

\begin{figure}[h]
   \centering    
   \includegraphics[width=3in]{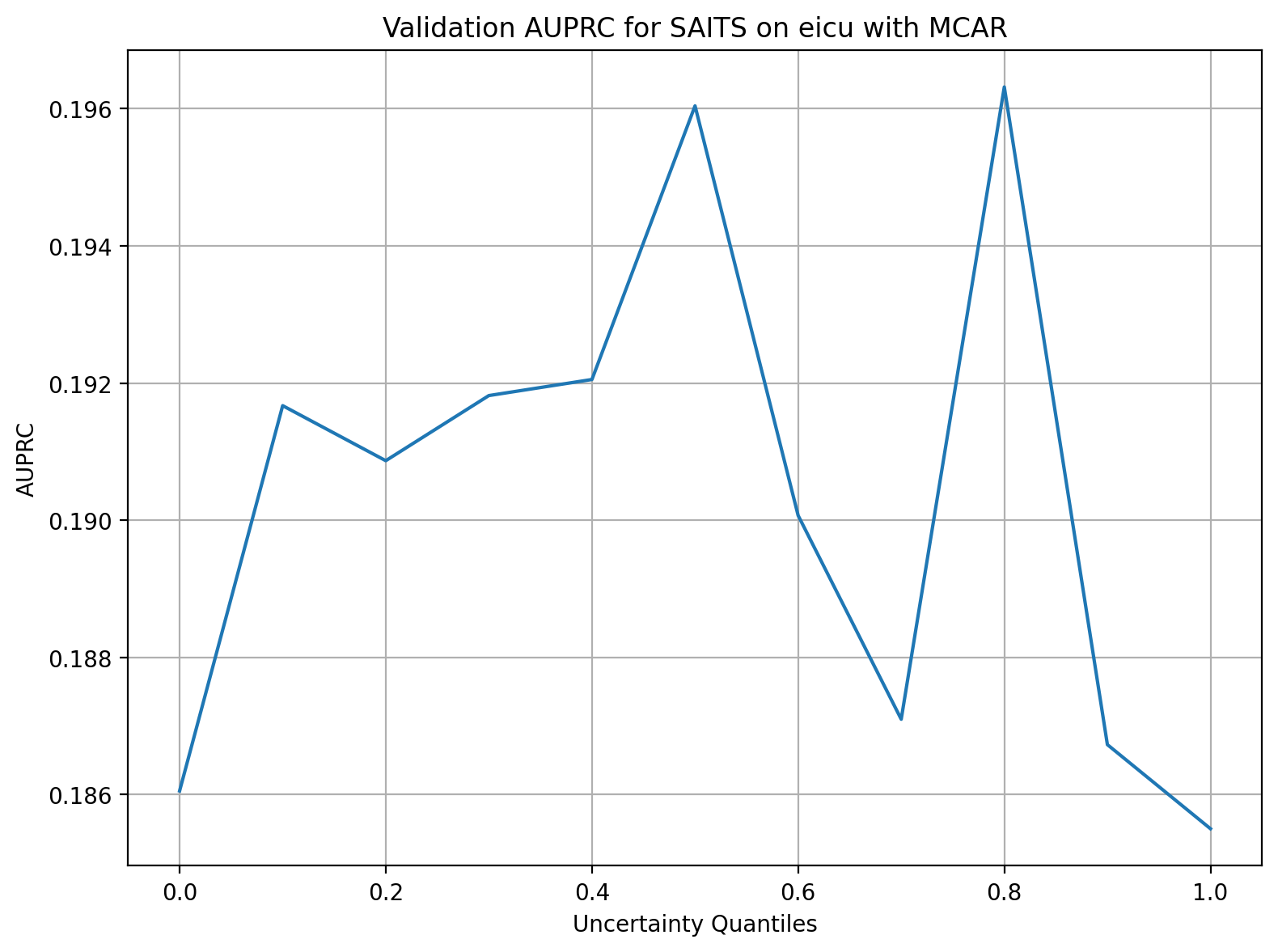} 
   \caption{Impact of Uncertainty on AUPRC for a classifier trained for mortality prediction, using a pretrained imputation model trained for MCAR missingness on eicu}
   \label{fig:fig:classfcn_eicu_SAITS_MCAR} 
 \end{figure}

\begin{figure}[h]
   \centering    
   \includegraphics[width=3in]{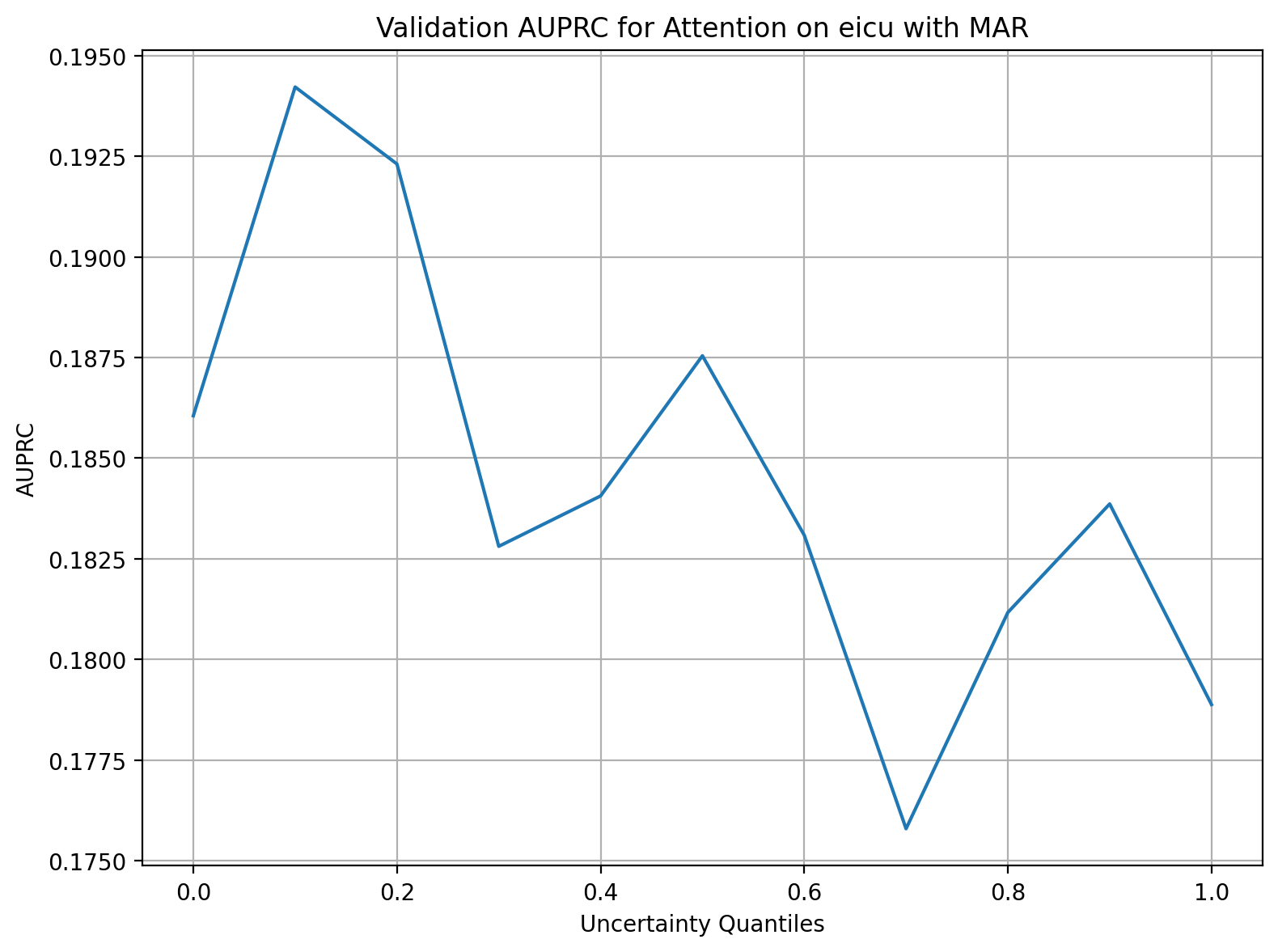} 
   \caption{Impact of Uncertainty on AUPRC for a classifier trained for mortality prediction, using a pretrained imputation model trained for MAR missingness on eicu}
   \label{fig:fig:classfcn_eicu_Attention_MAR} 
 \end{figure}

\begin{figure}[h]
   \centering    
   \includegraphics[width=3in]{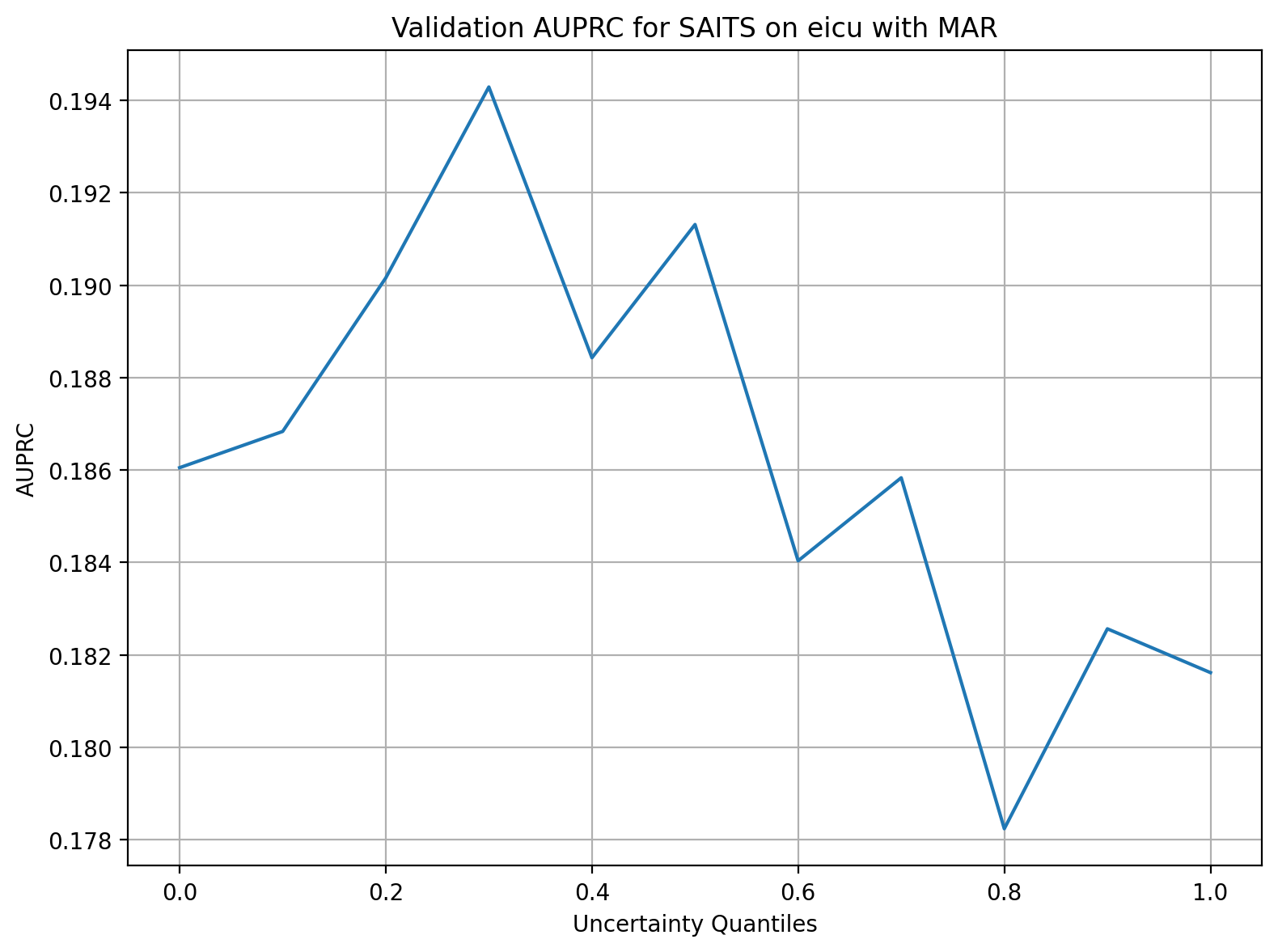} 
   \caption{Impact of Uncertainty on AUPRC for a classifier trained for mortality prediction, using a pretrained imputation model trained for MAR missingness on eicu}
   \label{fig:fig:classfcn_eicu_SAITS_MAR} 
 \end{figure}

\begin{figure}[h]
   \centering    
   \includegraphics[width=3in]{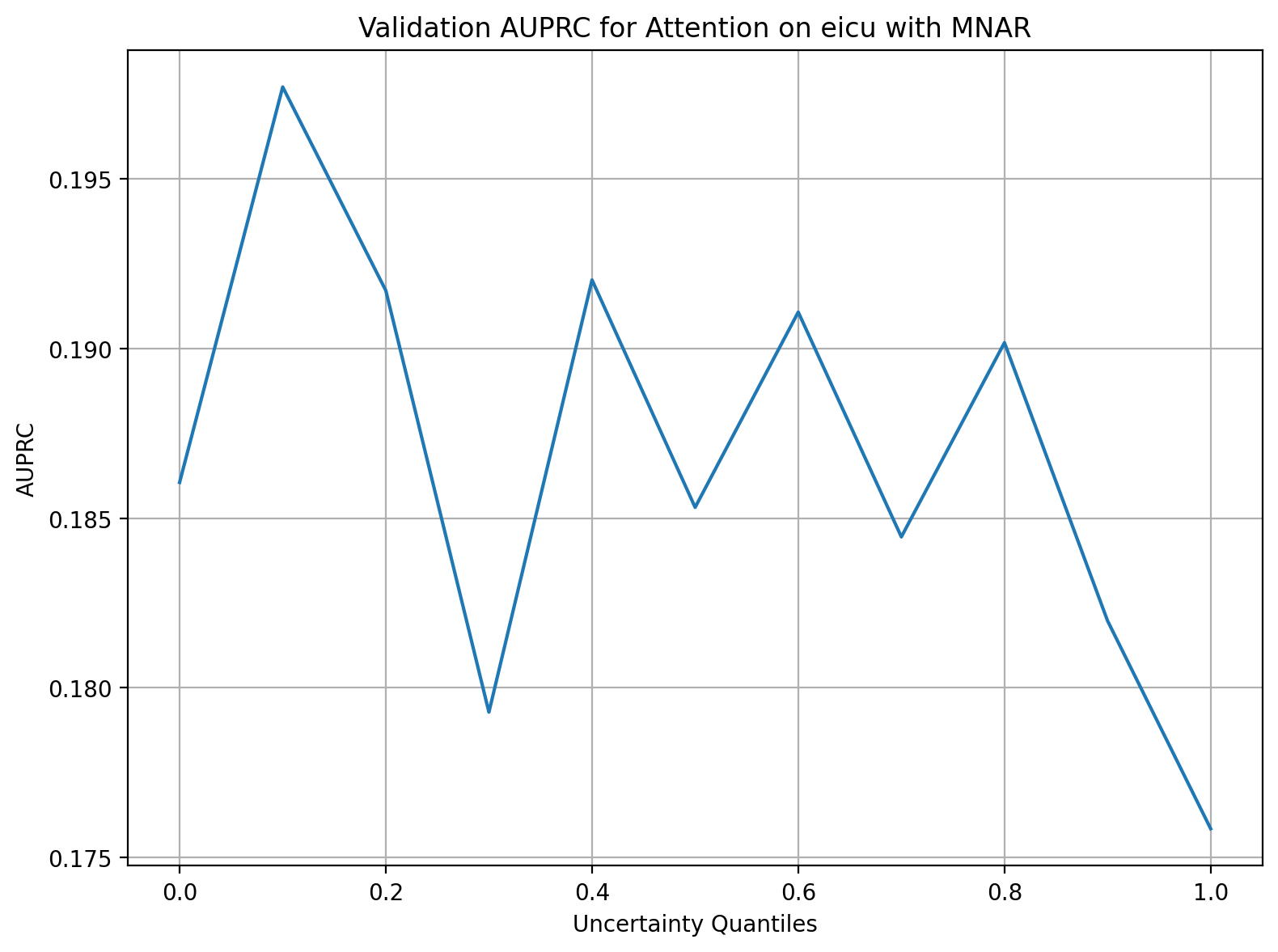} 
   \caption{Impact of Uncertainty on AUPRC for a classifier trained for mortality prediction, using a pretrained imputation model trained for MNAR missingness on eicu}
   \label{fig:fig:classfcn_eicu_Attention_MNAR} 
 \end{figure}

\begin{figure}[h]
   \centering    
   \includegraphics[width=3in]{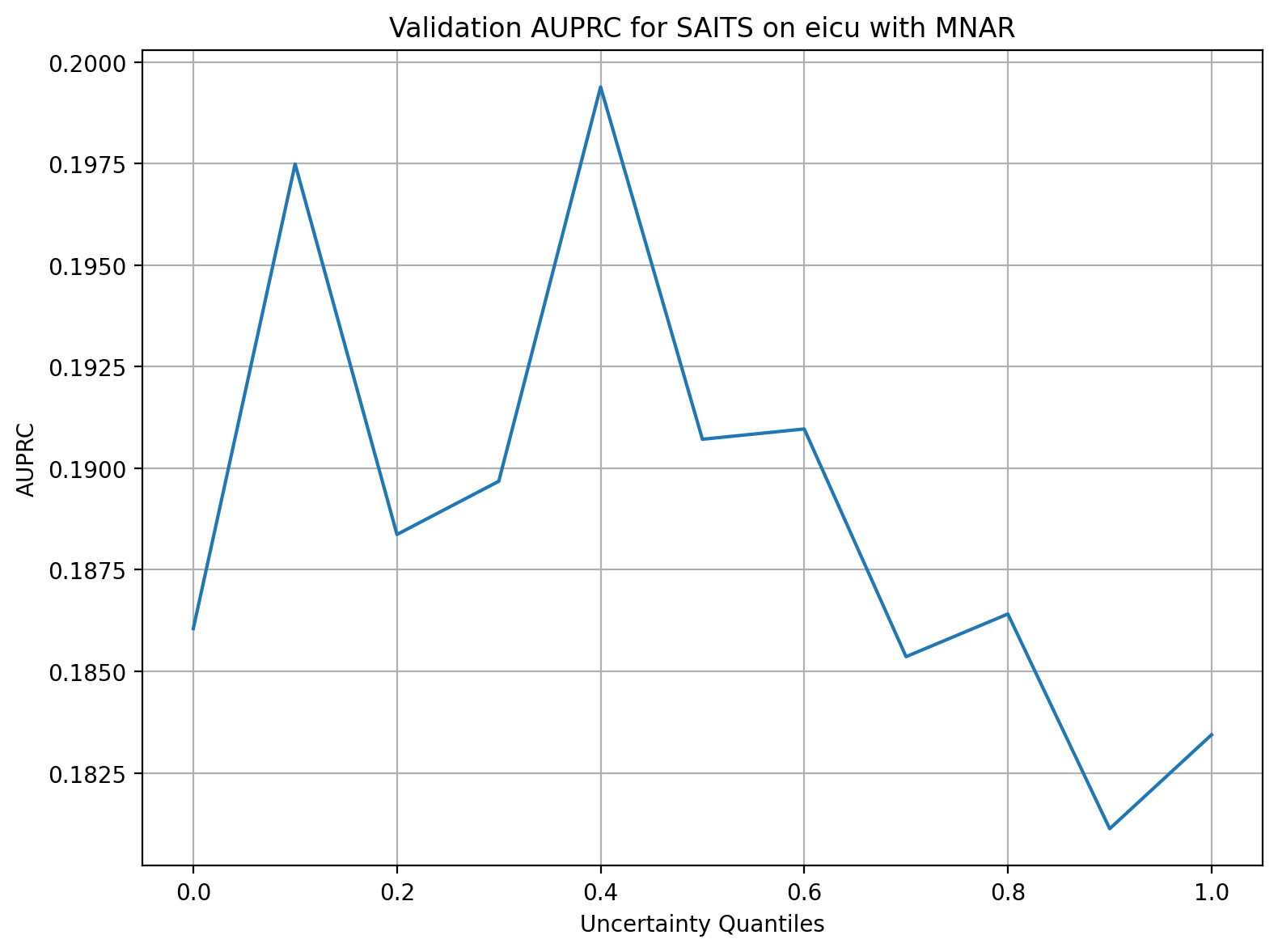} 
   \caption{Impact of Uncertainty on AUPRC for a classifier trained for mortality prediction, using a pretrained imputation model trained for MNAR missingness on eicu}
   \label{fig:fig:classfcn_eicu_SAITS_MNAR} 
 \end{figure}

\begin{figure}[h]
   \centering    
   \includegraphics[width=3in]{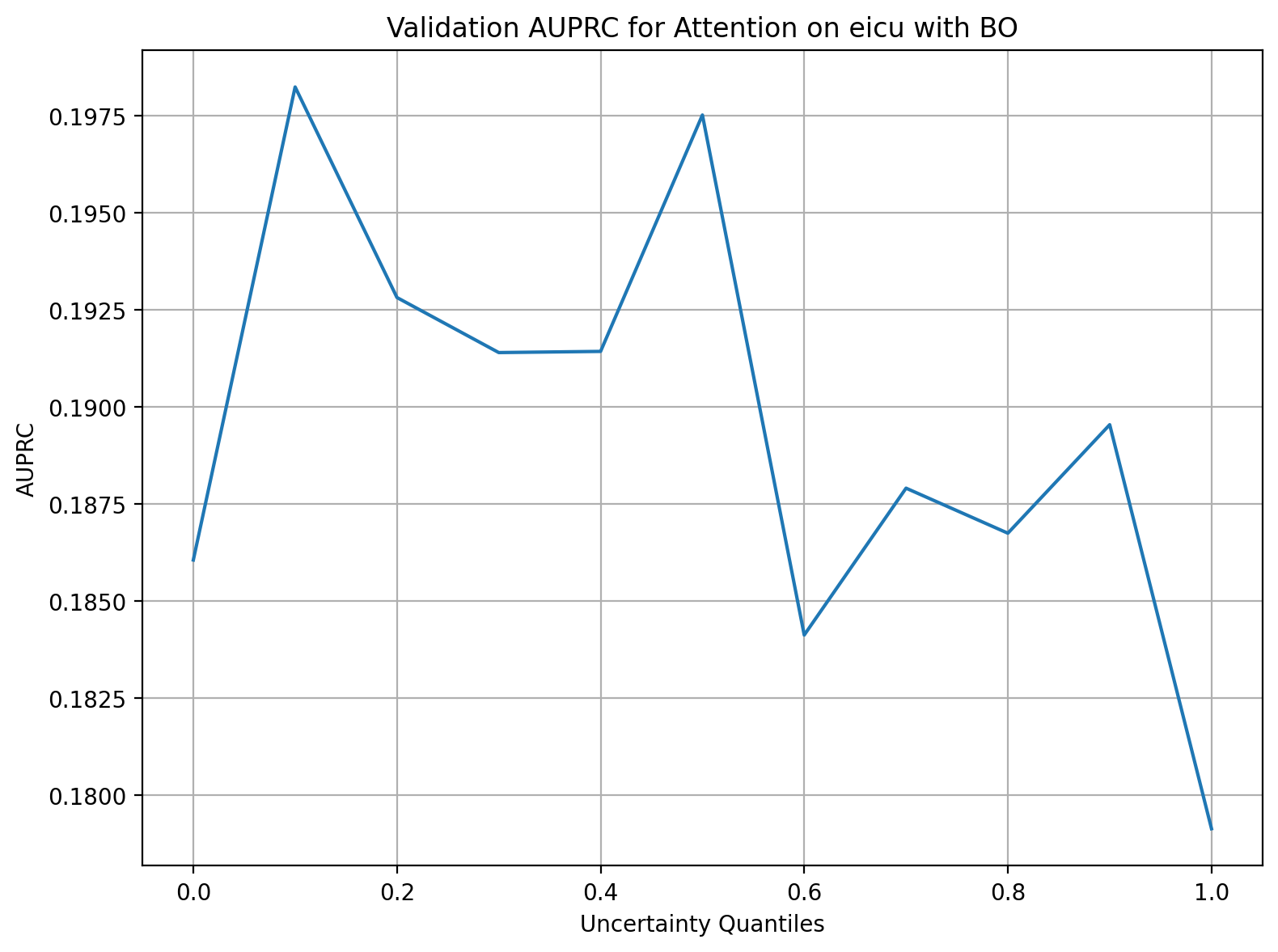} 
   \caption{Impact of Uncertainty on AUPRC for a classifier trained for mortality prediction, using a pretrained imputation model trained for BO missingness on eicu}
   \label{fig:fig:classfcn_eicu_Attention_BO} 
 \end{figure}

\begin{figure}[h]
   \centering    
   \includegraphics[width=3in]{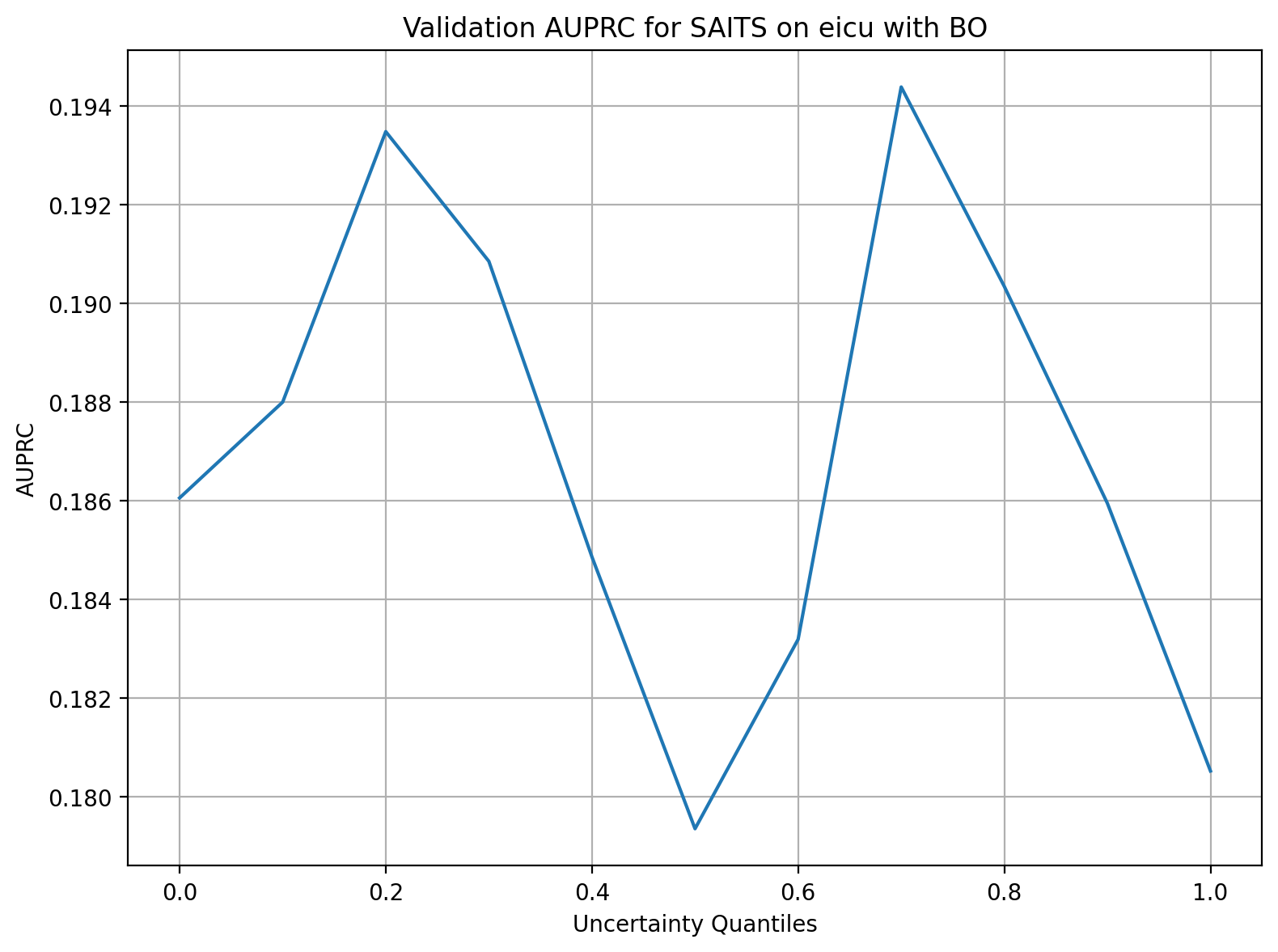} 
   \caption{Impact of Uncertainty on AUPRC for a classifier trained for mortality prediction, using a pretrained imputation model trained for BO missingness on eicu}
   \label{fig:fig:classfcn_eicu_SAITS_BO} 
 \end{figure}

\begin{figure}[h]
   \centering    
   \includegraphics[width=3in]{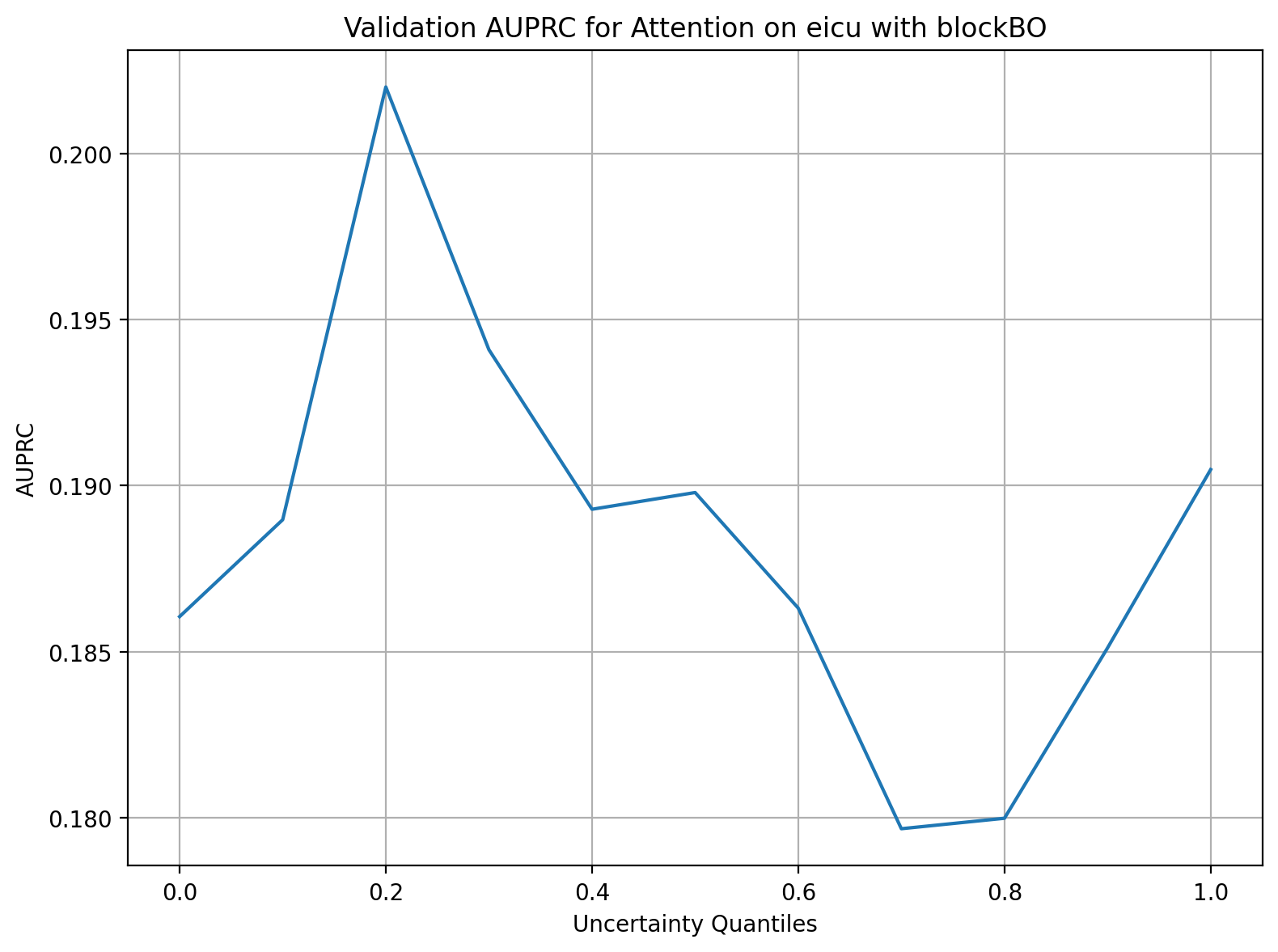} 
   \caption{Impact of Uncertainty on AUPRC for a classifier trained for mortality prediction, using a pretrained imputation model trained for blockBO missingness on eicu}
   \label{fig:fig:classfcn_eicu_Attention_blockBO} 
 \end{figure}

\begin{figure}[h]
   \centering    
   \includegraphics[width=3in]{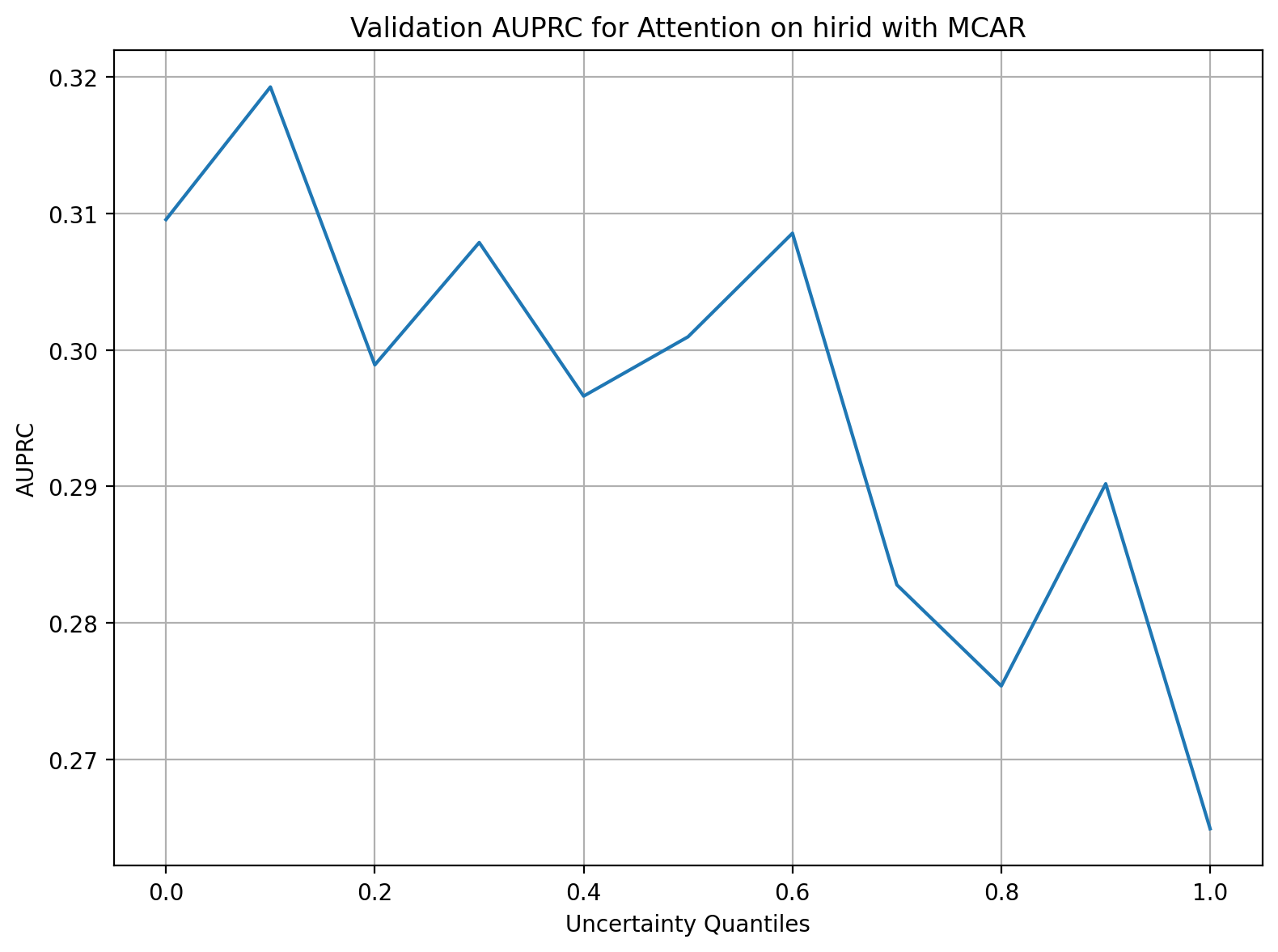} 
   \caption{Impact of Uncertainty on AUPRC for a classifier trained for mortality prediction, using a pretrained imputation model trained for MCAR missingness on hirid}
   \label{fig:fig:classfcn_hirid_Attention_MCAR} 
 \end{figure}

\begin{figure}[h]
   \centering    
   \includegraphics[width=3in]{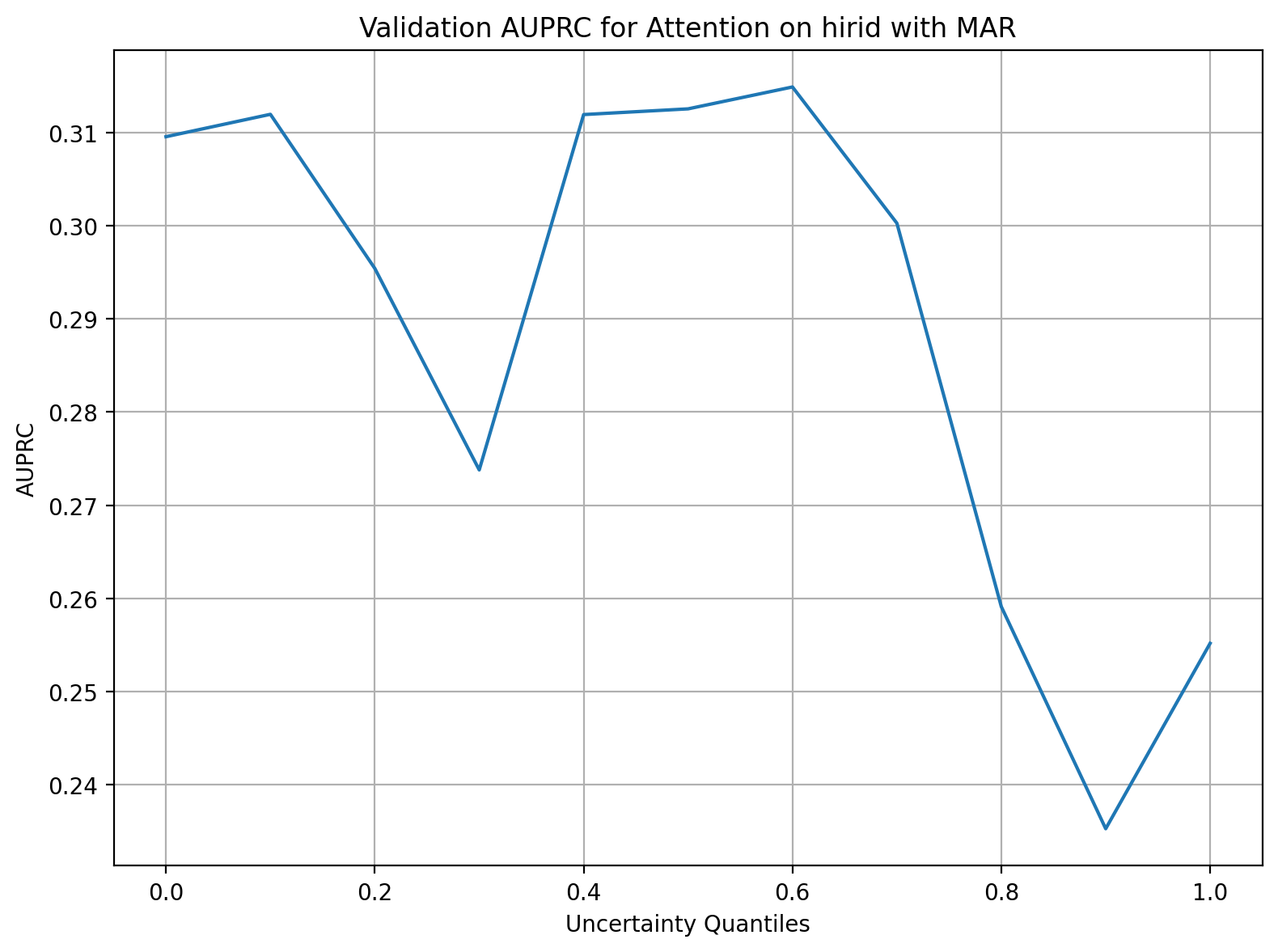} 
   \caption{Impact of Uncertainty on AUPRC for a classifier trained for mortality prediction, using a pretrained imputation model trained for MAR missingness on hirid}
   \label{fig:fig:classfcn_hirid_Attention_MAR} 
 \end{figure}

\begin{figure}[h]
   \centering    
   \includegraphics[width=3in]{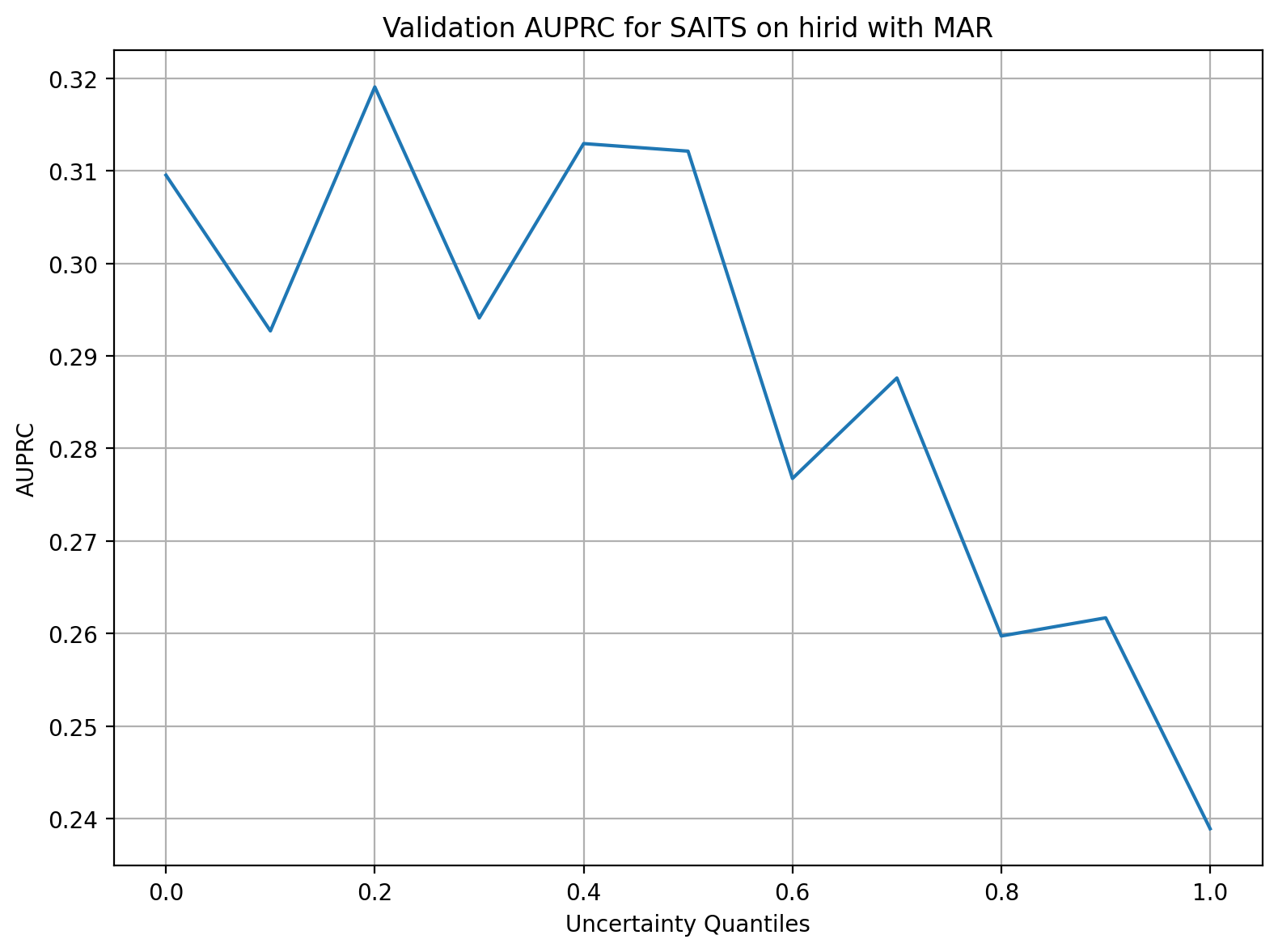} 
   \caption{Impact of Uncertainty on AUPRC for a classifier trained for mortality prediction, using a pretrained imputation model trained for MAR missingness on hirid}
   \label{fig:fig:classfcn_hirid_SAITS_MAR} 
 \end{figure}

\begin{figure}[h]
   \centering    
   \includegraphics[width=3in]{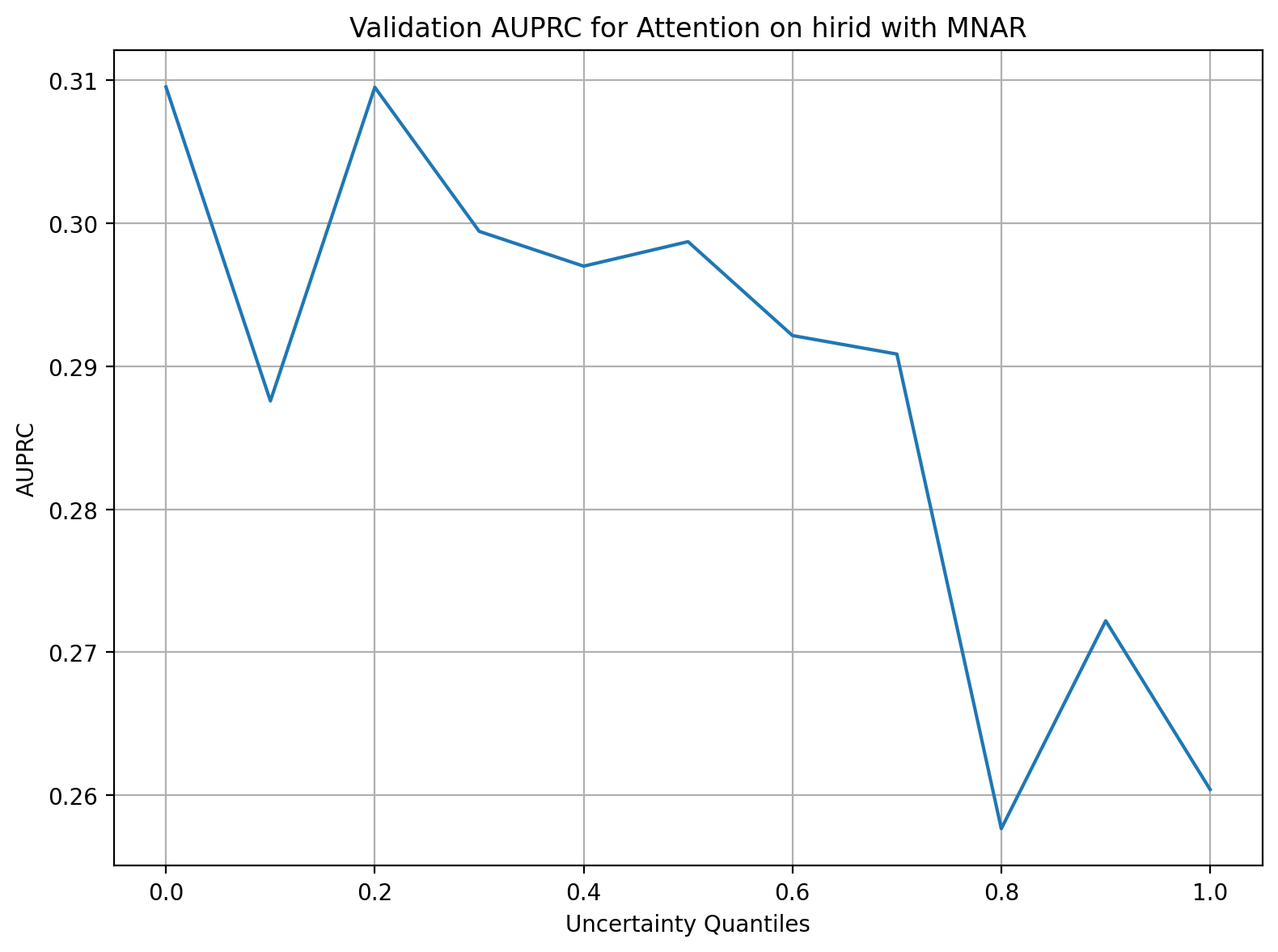} 
   \caption{Impact of Uncertainty on AUPRC for a classifier trained for mortality prediction, using a pretrained imputation model trained for MNAR missingness on hirid}
   \label{fig:fig:classfcn_hirid_Attention_MNAR} 
 \end{figure}

\begin{figure}[h]
   \centering    
   \includegraphics[width=3in]{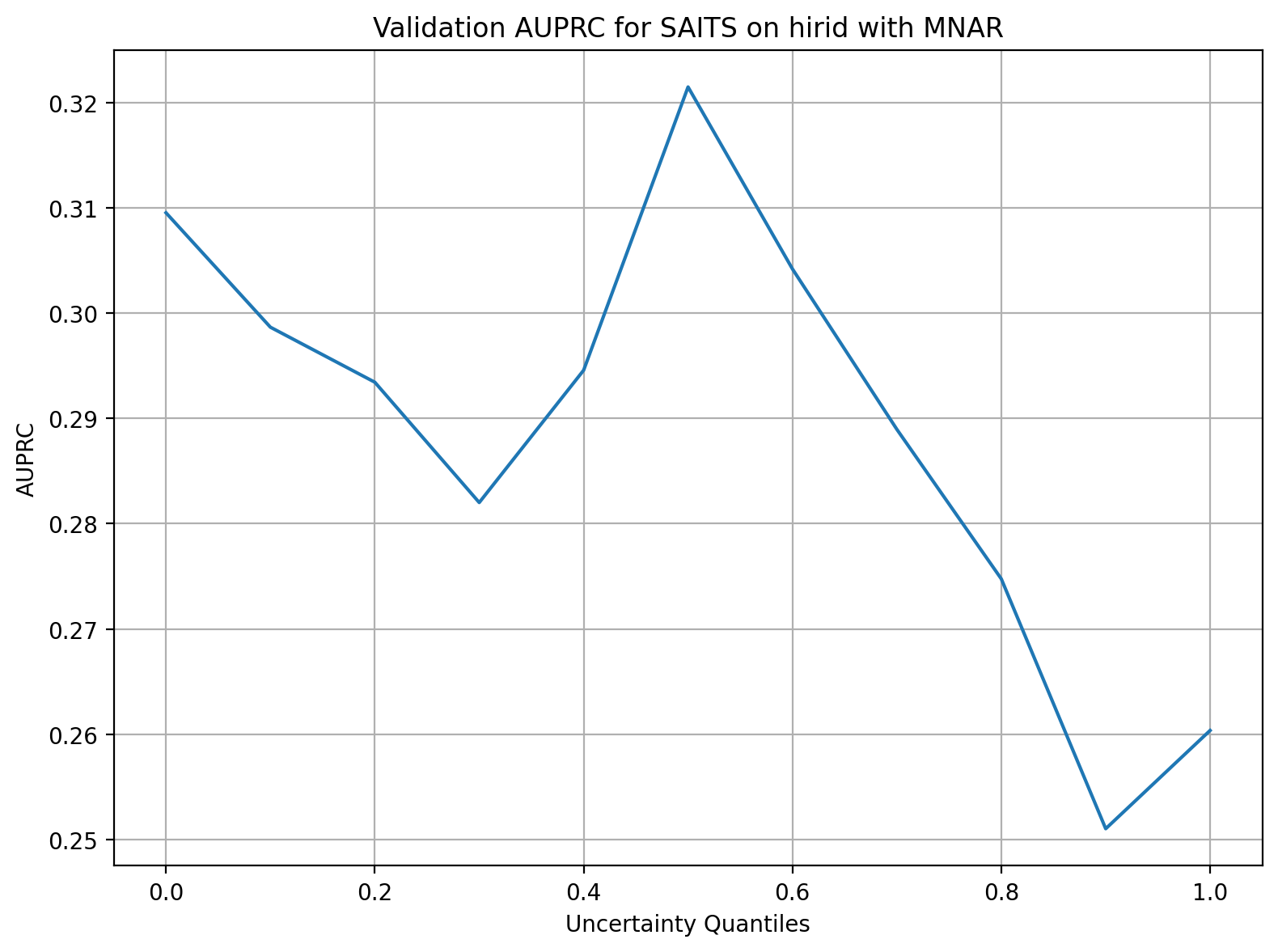} 
   \caption{Impact of Uncertainty on AUPRC for a classifier trained for mortality prediction, using a pretrained imputation model trained for MNAR missingness on hirid}
   \label{fig:fig:classfcn_hirid_SAITS_MNAR} 
 \end{figure}

\begin{figure}[h]
   \centering    
   \includegraphics[width=3in]{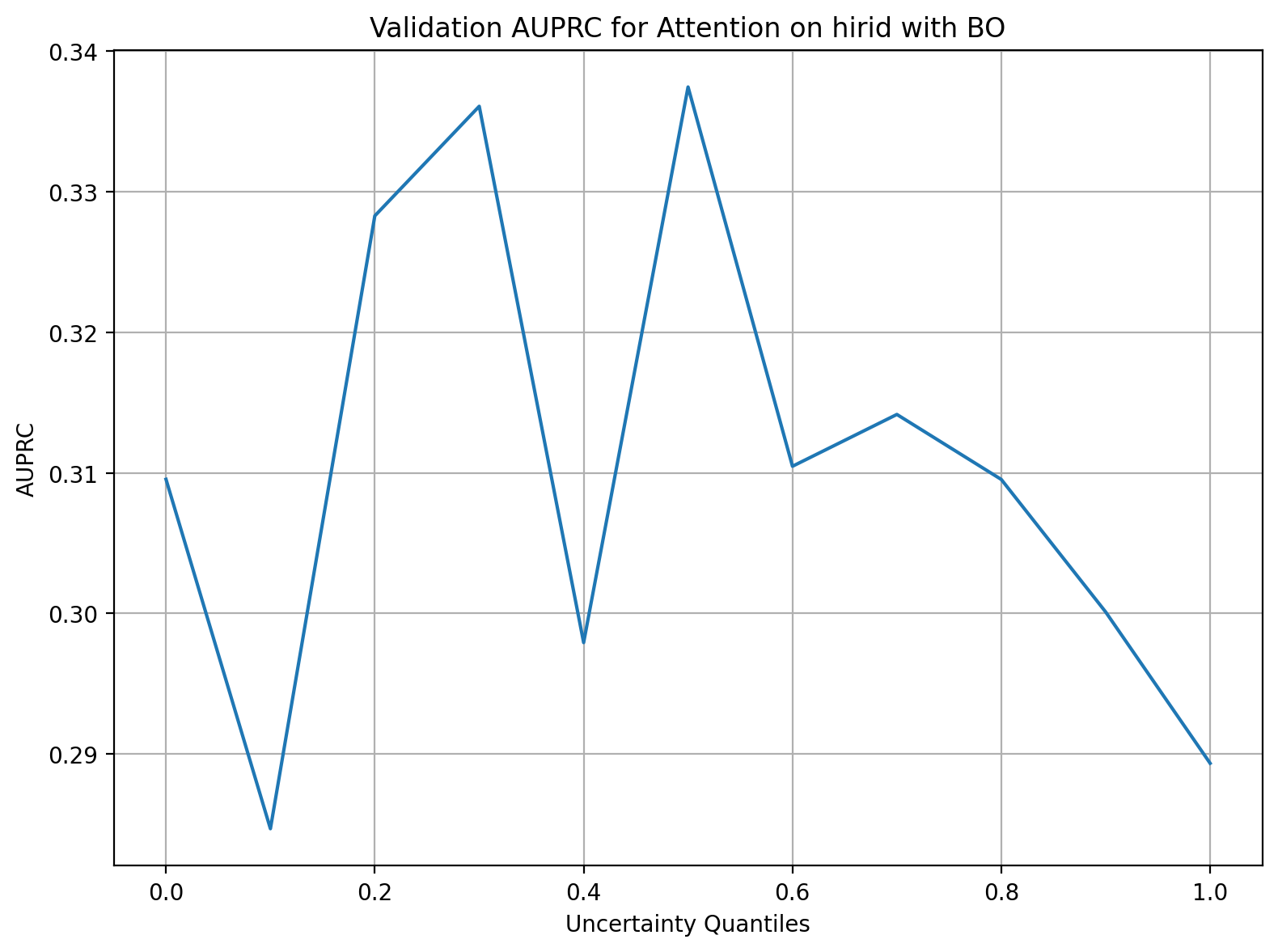} 
   \caption{Impact of Uncertainty on AUPRC for a classifier trained for mortality prediction, using a pretrained imputation model trained for BO missingness on hirid}
   \label{fig:fig:classfcn_hirid_Attention_BO} 
 \end{figure}

\begin{figure}[h]
   \centering    
   \includegraphics[width=3in]{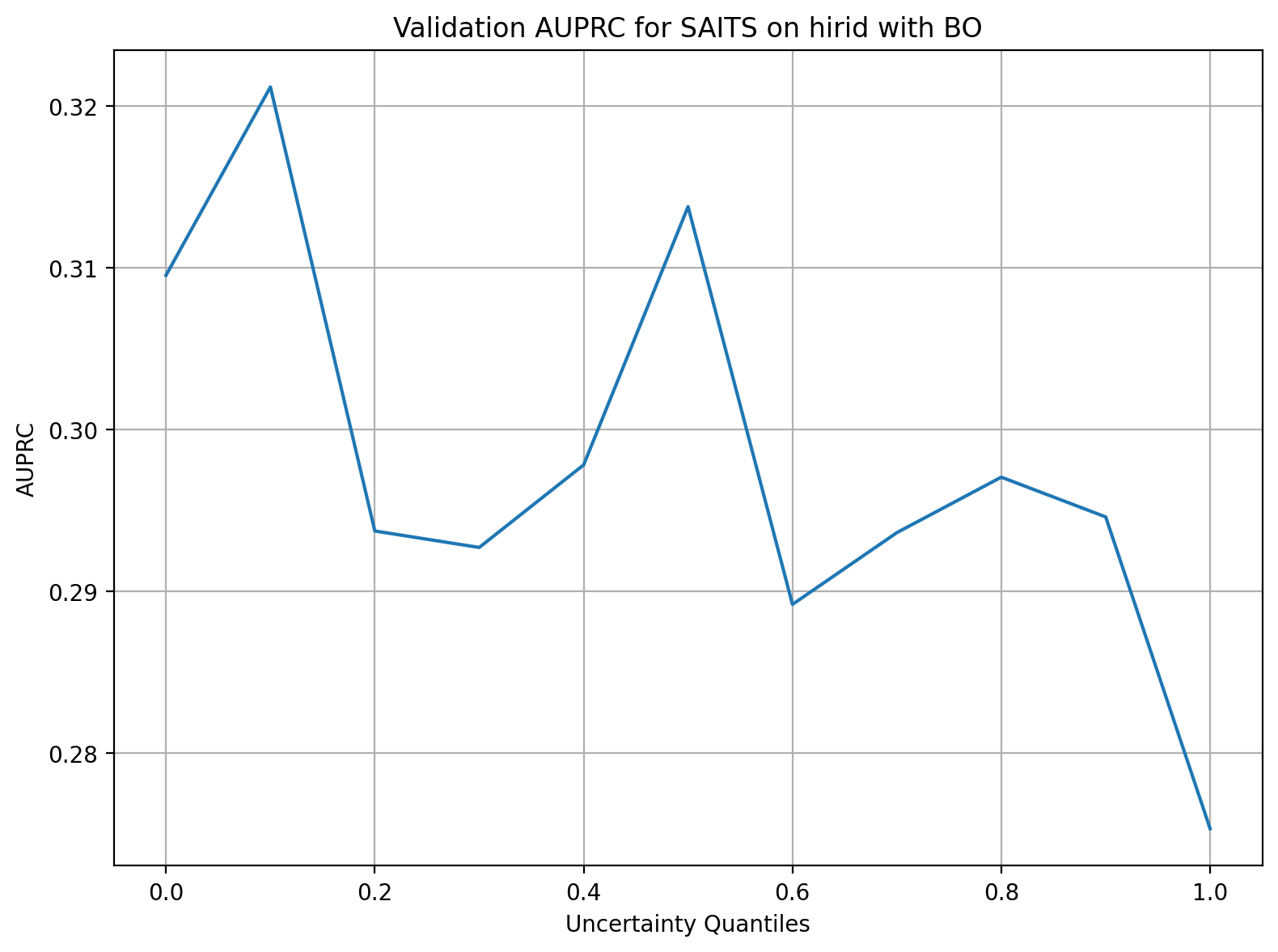} 
   \caption{Impact of Uncertainty on AUPRC for a classifier trained for mortality prediction, using a pretrained imputation model trained for BO missingness on hirid}
   \label{fig:fig:classfcn_hirid_SAITS_BO} 
 \end{figure}

\begin{figure}[h]
   \centering    
   \includegraphics[width=3in]{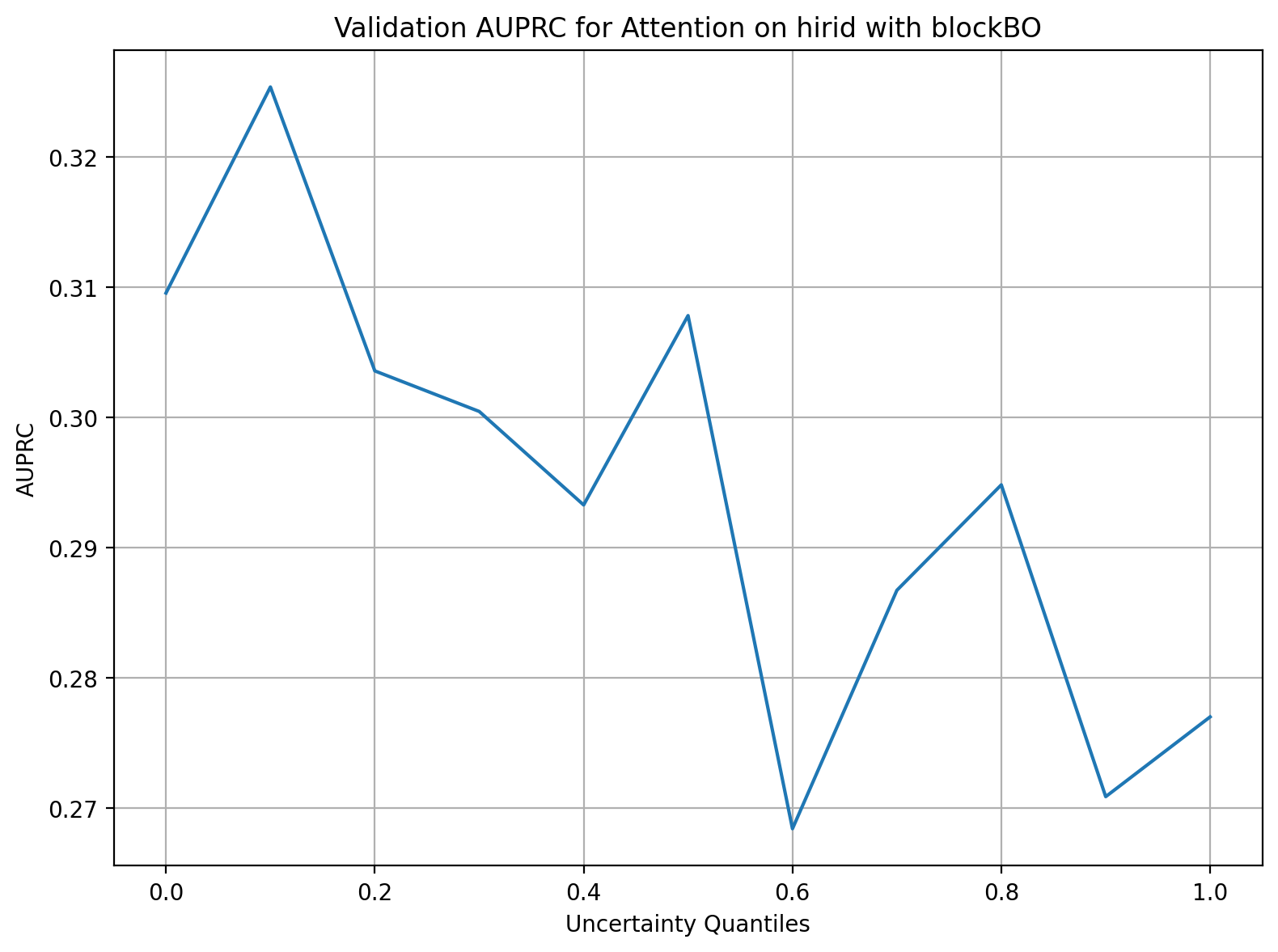} 
   \caption{Impact of Uncertainty on AUPRC for a classifier trained for mortality prediction, using a pretrained imputation model trained for blockBO missingness on hirid}
   \label{fig:fig:classfcn_hirid_Attention_blockBO} 
 \end{figure}

\begin{figure}[h]
   \centering    
   \includegraphics[width=3in]{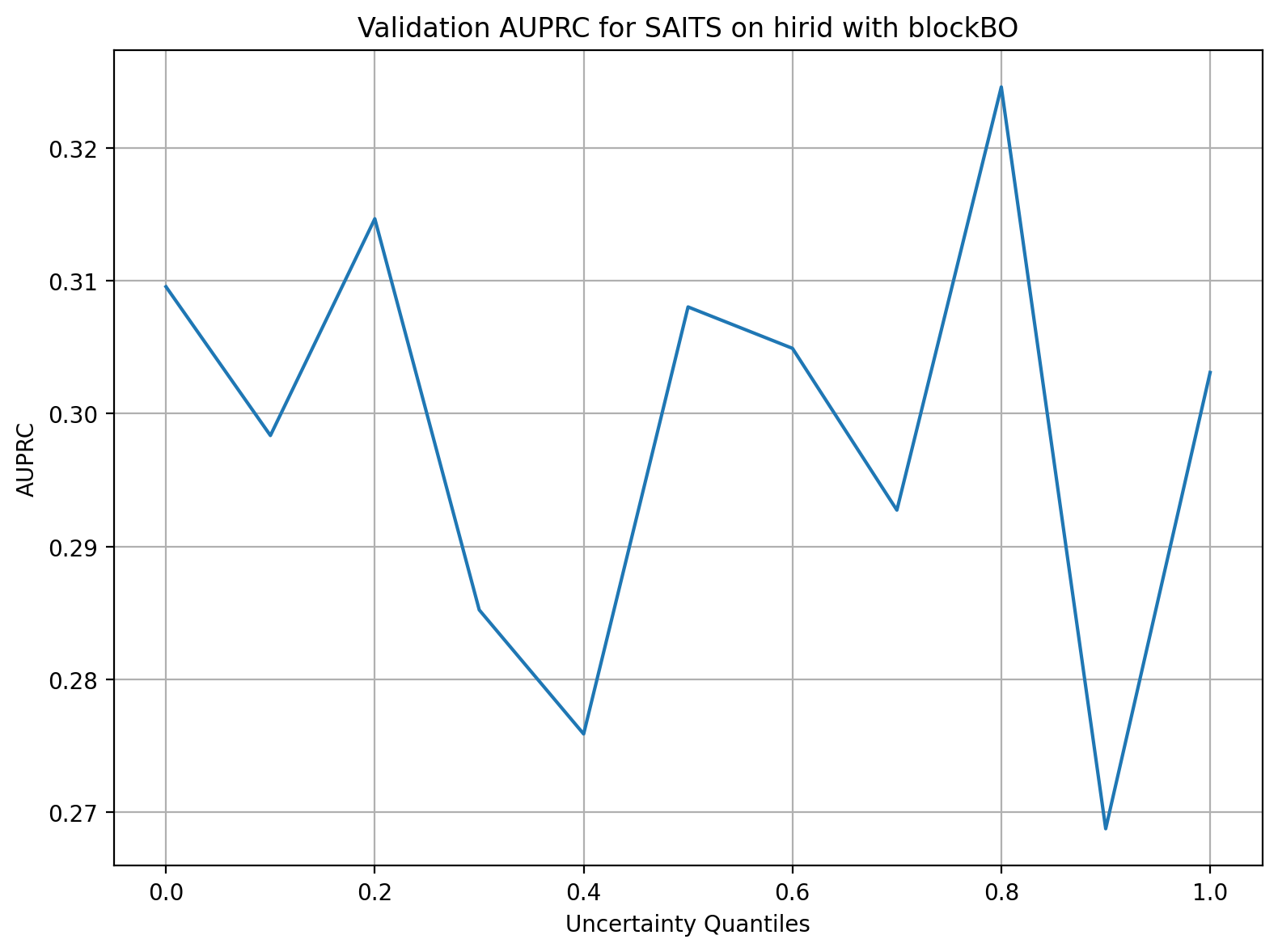} 
   \caption{Impact of Uncertainty on AUPRC for a classifier trained for mortality prediction, using a pretrained imputation model trained for blockBO missingness on hirid}
   \label{fig:fig:classfcn_hirid_SAITS_blockBO} 
 \end{figure}

\FloatBarrier
\newpage
\section{Baseline Comparisons}
\label{AppendixB}

\begin{figure}[H]
   \centering    
   \includegraphics[width=6in]{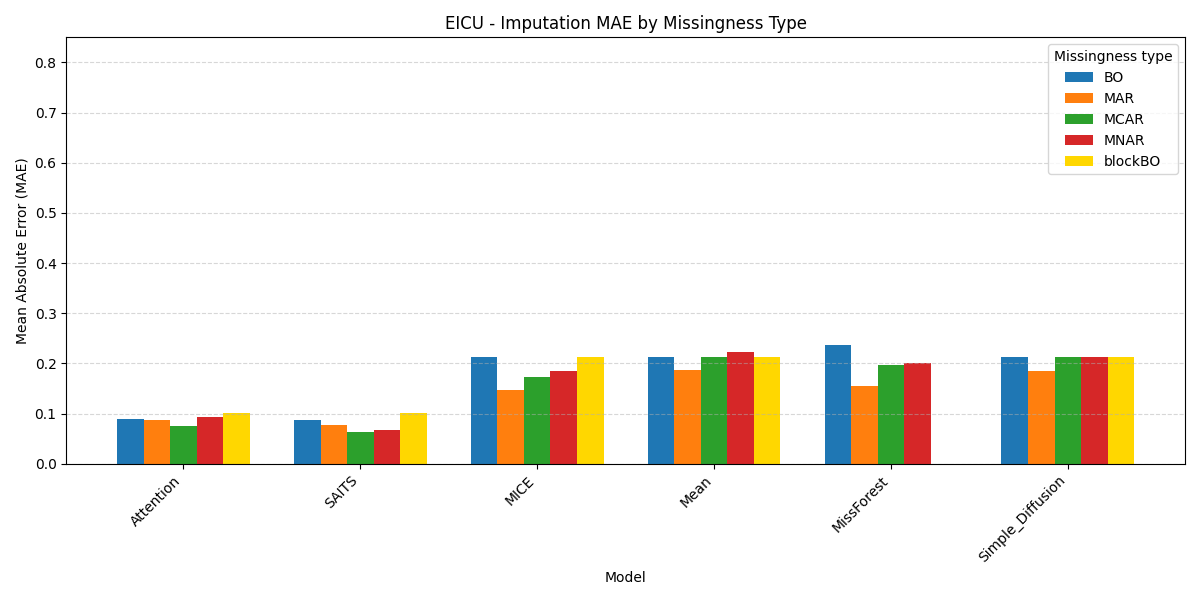} 
   \caption{Imputation baseline comparison on the eicu dataset. Note that the 'Attention' method is a vanilla Transformer encoder.}
   \label{fig:eicu_imp_benchmark_plot} 
 \end{figure}

 \begin{figure}[H]
   \centering    
   \includegraphics[width=6in]{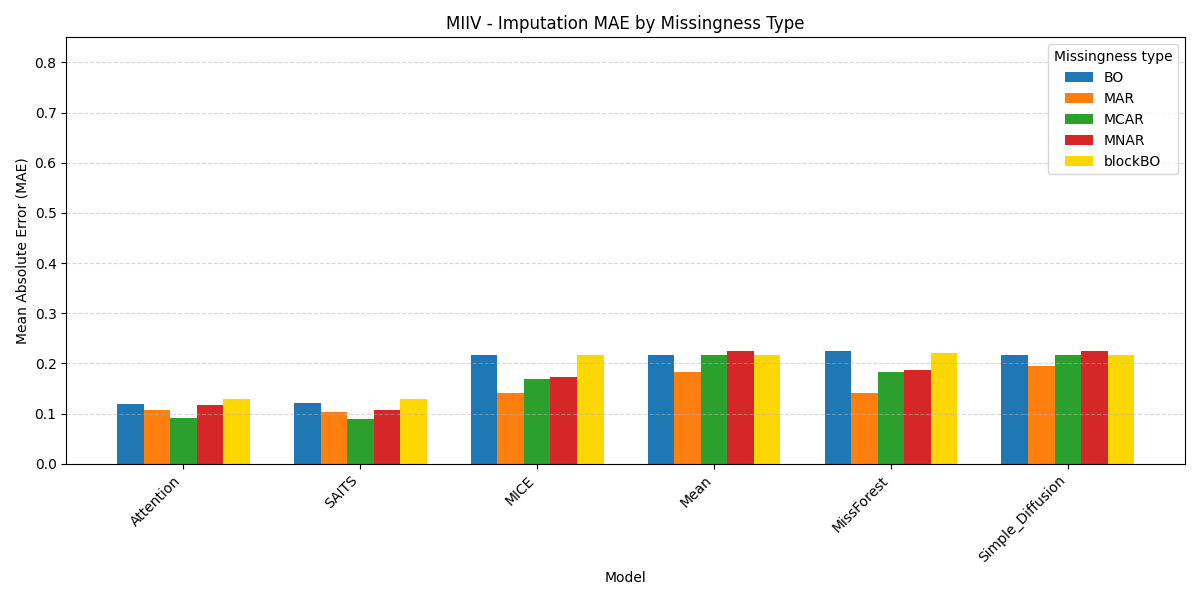} 
   \caption{Imputation baseline comparison on the MIMIC IV dataset. Note that the 'Attention' method is a vanilla Transformer encoder.}
   \label{fig:miiv_imp_benchmark_plot} 
 \end{figure}

 \begin{figure}[H]
   \centering    
   \includegraphics[width=6in]{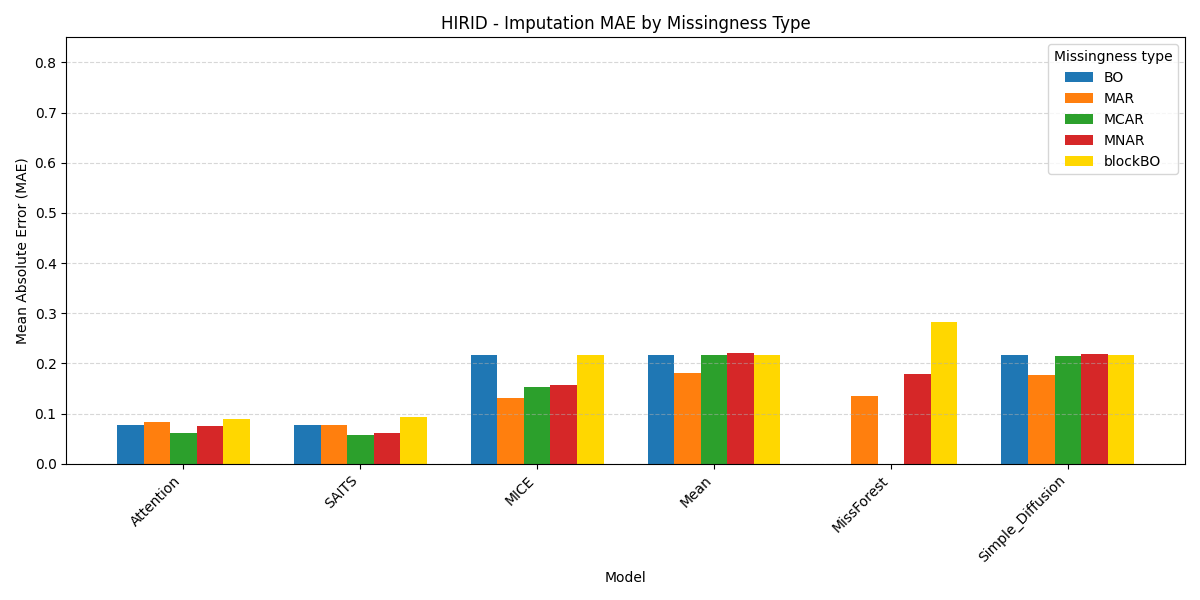} 
   \caption{Imputation baseline comparison on the HiRID dataset. Note that the 'Attention' method is a vanilla Transformer encoder.}
   \label{fig:hirid_imp_benchmark_plot} 
 \end{figure}

\end{document}